\def\year{2021}\relax
\documentclass[letterpaper]{article} %
\usepackage{aaai21}  %
\usepackage{times}  %
\usepackage{helvet} %
\usepackage{courier}  %
\usepackage[hyphens]{url}  %
\usepackage{graphicx} %
\usepackage{natbib}  %
\usepackage{caption} %
\title{Controllable Guarantees for Fair Outcomes via\\ Contrastive Information Estimation}
\author{
   Umang Gupta\textsuperscript{\rm 1},
   Aaron M Ferber\textsuperscript{\rm 2},
   Bistra Dilkina\textsuperscript{\rm 2},
   Greg Ver Steeg\textsuperscript{\rm 1}\\
}
\affiliations{
   \textsuperscript{\rm 1} Information Sciences Institute, University of Southern California \\
   \textsuperscript{\rm 2} University of Southern California \\
   \{umanggup, gregv\}@isi.edu, \{aferber, dilkina\}@usc.edu
}

\newif\iffull%
\fulltrue%
\newcommand{\full}[2]{%
\iffull%
   #1%
\else%
   #2%
\fi
}

\usepackage[ruled,vlined]{algorithm2e}
\usepackage[outercaption]{sidecap}
\usepackage{booktabs}
\usepackage{multirow}
\usepackage{tablefootnote}
\usepackage{tikz}
\usepackage{tabularx}
\usepackage{subcaption}
\usetikzlibrary{shapes,backgrounds}
\usetikzlibrary{patterns}
\usepackage{pifont}%
\newcommand{\cmark}{\ding{51}}%
\newcommand{\xmark}{\ding{55}}%
\usepackage{amsthm}

\iffull%
   \usepackage{color}
   \usepackage[hidelinks]{hyperref}
   \hypersetup{
      colorlinks = true,
      citecolor = blue,
   }
\fi

\graphicspath{ {./figures/} }

\usepackage{amsmath}
\usepackage{amsfonts}

\DeclareRobustCommand{\KL}[2]{\ensuremath{\textrm{KL}\left(#1\;\|\;#2\right)}}

\newcommand{\normal}{\mathcal{N}}

\newcommand{\real}{\mathbb R}
\newcommand{\E}{\mathbb E}

\DeclareMathOperator*{\argmax}{arg\,max}

\newcommand{\mba}{\mathbf{a}}
\newcommand{\mbb}{\mathbf{b}}
\newcommand{\mbc}{\mathbf{c}}

\newcommand{\mbu}{\mathbf{u}}
\newcommand{\mbv}{\mathbf{v}}
\newcommand{\mbw}{\mathbf{w}}
\newcommand{\mbx}{\mathbf{x}}
\newcommand{\mby}{\mathbf{y}}
\newcommand{\mbz}{\mathbf{z}}

\newcommand{\mbW}{\mathbf{W}}

\DeclareMathSymbol{:}{\mathord}{operators}{"3A}
\newcommand{\equal}{=}

\newcommand{\DP}{\ensuremath{\Delta_{\text{DP}}}}

\newcommand{\codeurl}{\url{https://github.com/umgupta/fairness-via-contrastive-estimation}}
\newcommand{\appendixurl}{\url{https://arxiv.org/abs/2101.04108}}

\newtheorem{theorem}{Theorem}[]
\newtheorem{definition}[theorem]{Definition}

\newtheorem{proposition}[theorem]{Proposition}

\newtheorem{fact}[theorem]{Fact}

\newtheorem*{theorem*}{Theorem}
\newtheorem*{definition*}{Definition}
\newtheorem*{assumption*}{Assumption}
\newtheorem*{conjecture*}{Conjecture}
\newtheorem*{claim*}{Claim}
\newtheorem*{lemma*}{Lemma}
\newtheorem*{proposition*}{Proposition}
\newtheorem*{property*}{Property}
\newtheorem*{fact*}{Fact}
\newtheorem*{corollary*}{Corollary}
\newtheorem*{example*}{Example}
\newtheorem*{remark*}{Remark}
\newtheorem*{exercise*}{Exercise}

\begin{document}
\maketitle

\begin{abstract}

    Controlling bias in training datasets is vital for ensuring equal treatment, or parity, between different groups in downstream applications. A naive solution is to transform the data so that it is statistically independent of group membership, but this may throw away too much information when a reasonable compromise between fairness and accuracy is desired. Another common approach is to limit the ability of a particular adversary who seeks to maximize parity. Unfortunately, representations produced by adversarial approaches may still retain biases as their efficacy is tied to the complexity of the adversary used during training. To this end, we theoretically establish that by limiting the mutual information between representations and protected attributes, we can assuredly control the parity of any downstream classifier. We demonstrate an effective method for controlling parity through mutual information based on contrastive information estimators and show that they outperform approaches that rely on variational bounds based on complex generative models. We test our approach on \textit{UCI Adult} and \textit{Heritage Health} datasets and demonstrate that our approach provides more informative representations across a range of desired parity thresholds while providing strong theoretical guarantees on the parity of any downstream algorithm.

\end{abstract}

\section{Introduction}\label{sec:intro}
    Decisions based on biased data can promote biased outcomes.
    Learning algorithms often exploit and exaggerate biases present in the training dataset.
    One way to prevent algorithms from reproducing bias in the data is to pre-process the data so that information about protected attributes is removed~\cite{song2019learning,madras2018learning,McNamara}.
    Ideally, users of this transformed data can focus on maximizing performance for their tasks using any available method without the risk of producing unfair outcomes~\cite{cissefairness}.
    The strongest requirement for fair representations is to be statistically independent of sensitive attributes, but this may lead to large drops in predictive performance as sensitive attributes are often correlated with the target. Therefore, it is desirable to produce representations that can trade-off
    some measure of fairness (e.g., statistical parity in this work) with utility~\cite{KrishnaMenon2018,Dutta}.%

    Many recent approaches for learning fair representations leverage adversarial learning to remove unwanted biases from the data by limiting the ability of an adversary  to reconstruct sensitive attributes from the representation during training~\cite{jaiswal2020invariant,roy2019mitigating, song2019learning,madras2018learning,edwards2015censoring}.
    While adversarial methods have been shown to be useful in learning fair representations, they are often limited by the adversary's model capacity. A more powerful adversary than the one used during training may reveal hidden biases that are present in the representations~\cite{xu2020theory}.
    As a result, a model trained to control fairness against one adversary has no guarantee to control fairness against an arbitrary adversary.  %
    Other methods for learning fair representations focus on inducing statistical independence~\cite{moyer2018invariant, Louizos2015}.
    While fairness guarantees of these methods are agnostic to the downstream algorithms, including adversarial attempts to exploit bias, they are inefficient at trading-off between fairness and informativeness in the representation, which is often desirable for fairness applications.
    Some methods rely on complex generative modeling of  data to discover representations that are invariant to the protected attributes while preserving as much information as possible about the data. However, the quality of the generative model is a performance bottleneck. We summarize some of these approaches and their properties in Table~\ref{tab:comparison}.

    \newcommand{\strong}{\textbf{Strong}}
    \newcommand{\weak}{Weak}
    \newcommand{\no}{No}
    \newcommand{\none}{None}
    \newcommand{\heuristic}{Heuristic}
    \newcommand{\provable}{\textbf{Provable}}

    \begin{table}[t!]
            \centering
            \begin{tabular}{lcc}
                \toprule
                \multirow{2}{*}{\centering Representative Methods}
                                & Adversarial    & Controllable   \\
                                & Guarantee         & Parity         \\
                \cmidrule(r){1-1} \cmidrule(lr){2-3}
                \citet{song2019learning}   & \weak\tablefootnote{\citet{song2019learning} minimize two different bounds on $I(\mbz:\mbc)$ --- one is a very loose upper bound, and another uses adversarial learning. So the adversarial guarantee is unclear or, at best weaker.} & \heuristic \\
                \citet{moyer2018invariant} & \strong    & \no\tablefootnote{\citet{moyer2018invariant} designed their method for enforcing independence; however, we consider a modification of their method for controlling parity, based on our Thm.~\ref{thm:parity_mi_relation}.}\\
                \citet{madras2018learning} & \none    & \heuristic  \\
                \citet{roy2019mitigating}               & \none    & No \\
                Ours                                    & \strong  & \provable  \\
                \bottomrule
            \end{tabular}
            \caption{Fair representation learning methods}
            \label{tab:comparison}
    \end{table}

    In this paper, we focus on a widely used fairness measure known as statistical parity, though other measures may be more appropriate for specific problems.
    The difference in outcomes for two groups, or parity, is often used to quantify the fairness of a decision procedure. It has also been shown to best capture people's perception of fairness~\cite{srivastava2019}.  One of the main objectives of fair representation learning is to limit the parity of any possible downstream decision algorithm.  To this end, we relate the parity of any possible decision algorithm to an algorithm-agnostic measure of dependence. We show that we can provably control the parity of any possible decision algorithm that acts only on these representations by limiting  mutual information between the representations and the sensitive attributes.

    Estimating and bounding mutual information from data is a challenging and active field of research~\cite{poole2019variational}.
    We propose practical ways to limit the mutual information via contrastive estimates of conditional mutual information, thereby bypassing the need to use complex generative models that require explicit assumptions about the input distribution.
    Contrastive information measures have demonstrated state-of-the-art performance on many representation learning tasks~\cite{cpc,dgim,mikolov2013distributed} but not for fair representation learning.
    We evaluate our approach on two fairness benchmark datasets --- \textit{UCI Adult} and \textit{Heritage Health} dataset and show that representations provided by our method preserve more information at the desired fairness threshold compared to other adversarial as well as non-adversarial baselines.

    Our main contributions are --- a) We theoretically show that  mutual information between the representations and the sensitive attributes bounds the statistical parity of any decision algorithm; b) We propose practical ways to limit mutual information leveraging contrastive information estimators to efficiently trade-off predictability and accuracy.

\section{Mutual Information Bounds Parity}\label{sec:mi_parity}

    We consider a dataset of triplets $\mathcal D = \{x_i,y_i,c_i\}_{i=1}^N$, where $x_i,y_i,c_i$ are iid samples from data distribution $p(\mbx,\mby,\mbc)$.
    $\mbc$ are the sensitive or protected attributes, $\mby$ is the label, $\mbx$ are features of the sample which may include sensitive attributes
    and $\hat \mby$ denotes predicted label, according to some algorithm. We may also interpret $\hat \mby$ as the outcome of some decision procedure.
    We use bold letters to denote the random variable, and  the regular font represents corresponding samples.
    In this work, we consider stochastic representations of data i.e., $z(x)\sim q(\mbz|\mbx = x)$. We learn $d$-dimensional representations $z$ of input $x$, such that any classifier trained on only $z$ is guaranteed to be fair, i.e.,\  it has parity within some $\delta'$. In this work, we focus on statistical parity, a popular measure of group fairness, and it is defined as:

    \begin{definition}{\textbf{Statistical Parity:}}\label{def:parity}~\cite{Dwork2011}
        It is the absolute difference between the selection rates of two groups. Mathematically,
        \[\DP(\mathcal A, \mbc) = |P(\hat \mby = 1|\mbc=1) - P(\hat \mby=1|\mbc=0)|\]
        where $\hat \mby$ denotes decisions produced by some decision algorithm $\mathcal A$. {When there are more than two groups, we define statistical parity to be the maximum parity between any two groups (as implemented in \citet{Bird2020}).}
    \end{definition}

    \noindent
    Statistical parity is an algorithm dependent measure of fairness, whereas we require our representations to be fair for any downstream decision algorithm. Bearing this in mind, we show that  mutual information between the representations and the protected attributes, denoted as $I(\mbz:\mbc)$, can be used to limit the statistical parity via the following result.

    \begin{theorem}\label{thm:parity_mi_relation}
        For some $z, c\sim p(\mbz,\mbc)$, $z\in\real^d$, $c\in \{0,1\}$, and any decision algorithm $\mathcal A$ that acts on $z$, we have
        \[I(\mbz:\mbc)\geq g\left(\pi, \DP\left(\mathcal A,\mbc\right)\right)\] where $\pi=P(\mbc= 1)$ and  $g$ is a strictly increasing  non-negative convex function in $\DP\left(\mathcal A,\mbc\right)$.
    \end{theorem}
    \noindent
    We defer the proof of the above statement, the expression of $g$, and generalization of the theorem for multinomial  $\mbc$ to \full{Appendix~\ref{subsec:mi_parity_relation_proof}}{Appendix A.1}%
    \iffull\else\footnote{Appendix is available at \appendixurl.}\fi
    . We also visualize the bound on parity from Thm.~\ref{thm:parity_mi_relation} in \full{Appendix~\ref{subsec:mi_parity_relation_proof}}{Appendix A.1}.

    We know that if $I(\mbz:\mbc)=0$, then $\DP(\mathcal A, c)=0$, and by Thm.~\ref{thm:parity_mi_relation}, $g\left(\pi, \DP(\mathcal A, c)\right)$ will also be $0$. $I(\mbz:\mbc)$ upper bounds the function $g$.
    And since, $g$ is strictly increasing convex function in $ \DP(\mathcal A, c)$, $\DP(\mathcal A, c)$ will also be bounded.
    As a result, if $\mbz$ is a representation with bounded mutual information with $\mbc$, then any algorithm relying only on $\mbz$ to make a decision will also have bounded parity.
    Intuitively by reducing $I(\mbz:\mbc)$, we can  decrease \DP\ too.

    We remark that $I(\mbz:\mbc)$ has been  used as a proxy objective to control statistical parity previously~\cite{edwards2015censoring,song2019learning,moyer2018invariant}. It is often justified via the data processing inequality and the intuition that both statistical parity and mutual information are measures of dependence. However, using the data processing inequality, we can only guarantee that if we limit information about $\mbc$ in $\mbz$, then no subsequent operations on $\mbz$ can increase information about $\mbc$, i.e., $I(\hat \mby :\mbc) \leq I(\mbz:\mbc)$, but this fact alone implies nothing about statistical parity. Our result (Thm.~\ref{thm:parity_mi_relation}) demonstrates that limiting
    $I(\mbz:\mbc)$ will monotonically limit statistical parity, which had not been theoretically demonstrated until now.

\section{Practical Objectives for Controlling Parity}\label{sec:objective}
    Equipped with an algorithm agnostic upper bound to parity, we now discuss practical objectives for learning fair representations.
    Along with limiting parity, we also want the latent representation to be highly predictive (informative) about the label, which is often realized by maximizing mutual information between $\mby$ and $\mbz$,  i.e., $I(\mby:\mbz)$ implicitly~\cite{edwards2015censoring,madras2018learning,jaiswal2020invariant} or explicitly~\cite{moyer2018invariant}.
    \begin{align}
        \mathcal{O}_1 : &\max_q I(\mby:\mbz) \ \ \text{s.t. \ \ } I(\mbz:\mbc) \leq \delta \nonumber\\
    \text{or, } & \max_q I(\mby:\mbz) -\beta I(\mbz:\mbc) \label{eq:objective}
    \end{align}
    where, $\mbz, \mbx \sim q(\mbz|\mbx) p(\mbx)$ and $\beta>0$ is a hyperparameter.

    \subsection{Interference Between $I(\mby:\mbz)$ and $I(\mbz:\mbc)$}\label{subsec:interference}
        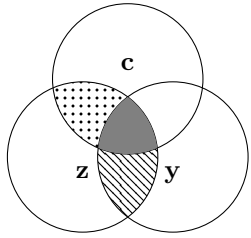
\begin{figure}
            \centering
            \def\firstcircle{(0,0) circle (1cm)}
            \def\secondcircle{(60:1.2cm) circle (1cm)}
            \def\thirdcircle{(0:1.2cm) circle (1cm)}
            \begin{tikzpicture}
                \begin{scope}[shift={(3cm,-5cm)}, fill opacity=1.0]
                    \firstcircle;
                    \secondcircle;
                    \thirdcircle;
                    \draw \firstcircle node[below] {$\mbz$};
                    \draw \secondcircle node[above] {$\mbc$};
                    \draw \thirdcircle node[below] {$\mby$};
                    \begin{scope}
                        \clip \firstcircle;
                        \fill[pattern=dots] \secondcircle;
                    \end{scope}
                    \begin{scope}
                        \clip \firstcircle;
                        \clip \thirdcircle;
                        \fill[pattern=north west lines] \thirdcircle;
                    \end{scope}
                    \begin{scope}
                        \clip \firstcircle;
                        \clip \secondcircle;
                        \fill[gray] \thirdcircle;
                    \end{scope}
                \end{scope}
            \end{tikzpicture}
            \caption{Venn diagram to show interference between $I(\mby:\mbz)$ and $I(\mbz:\mbc)$.
            See Sec.~\ref{subsec:interference} for details.}
            \label{fig:interference}
        \end{figure}

        While $I(\mby:\mbz)$ has been commonly proposed as a criterion to enforce the desiderata of representations being informative about labels, we argue that when the data is biased, i.e., $I(\mby:\mbc)>0$, maximizing $I (\mby :\mbz)$ is in direct contradiction with minimizing $I(\mbz:\mbc)$.
        To illustrate this point, we refer to the information Venn diagram in Fig.~\ref{fig:interference}. The goal of fair representation learning is to move the circle representing information about the representation, $\mbz$, to have high overlap with $\mby$ and low overlap with $\mbc$. However, there is a conflict in the gray region where we cannot increase overlap with $\mby$ without also increasing overlap with $\mbc$.
        In our experiments, we observe that this conflict hurts the model performance and makes it hard to achieve lower parity values at a reasonable accuracy (Fig.~\ref{fig:cond_vs_non_cond}).
        However, this conflict is not necessary. Since fair learning aims to capture information about $\mby$ that is \emph{not} related to the protected attribute $\mbc$, we want to maximize the overlap between $\mbz$ and the region of $\mby$ that excludes $\mbc$.
        This quantity is precisely the conditional mutual information, $I(\mby:\mbz|\mbc)$, which we propose to maximize instead of $I(\mby:\mbz)$. This leads us to the following  objective:
        \begin{align}
            \mathcal O_2: &
            \max_q I(\mby:\mbz|\mbc) \ \ \text{s.t. \ \ } I(\mbz:\mbc) \leq \delta
            \nonumber\\
            \text{or, } & \max_q I(\mby:\mbz|\mbc) -\beta I(\mbz:\mbc) \label{eq:conditional_objective}
        \end{align}
        where, $\mbz, \mbx \sim q(\mbz|\mbx) p(\mbx)$ and $\beta>0$ is a hyperparameter.
        Concurrent to this work,~\citet{rodriguez2020variational} have also proposed similar conditional mutual information term for retaining information about labels in the representation.
        Eq.~\ref{eq:conditional_objective} defines our approach, but the information-theoretic terms that appear are challenging to estimate directly.
        In the next sections, we derive practical variational bounds for these terms.%

        Before proceeding, we briefly discuss an ambiguity in the information-theoretic argument above.
        The triple overlap region of the information Venn diagram, sometimes called the  ``interaction information'', can be negative~\cite{beer}. This corresponds to the case that $\mbz$ and $\mbc$ have \emph{synergistic} information about $\mby$. That is, their combination allows us to predict $\mby$, even though individually they may not be predictive at all. The classic example of a synergistic interaction is the binary XOR relationship among three variables (example given in \full{Appendix~\ref{sec:synergy}}{Appendix B}). In that case, no variable can predict any other, but knowing any two variables perfectly predicts the third. If synergies are present, we may be able to attain a large value of $I(\mby:\mbz|\mbc)$ while $I(\mby:\mbz)=0$. In other words, maximizing $I(\mby:\mbz|\mbc)$ may include synergistic information, even if it hurts the ability to predict $\mby$ from $\mbz$ alone. While maximizing $I(\mby:\mbz)$ will not include synergistic information, it will prefer large positive interaction information. Positive interaction information can be interpreted as increasing the redundant information shared among $\mbz,\mby,\mbc$, even though this conflicts with our goal of excluding information about protected attributes.
        Different choices of objective will alter the prioritization of various multivariate information terms, but ultimately our choice is justified by the empirical performance shown in Sec.~\ref{sec:experiments}.

    \subsection{Lower Bound for $I(\mby:\mbz)$ and $I(\mby:\mbz|\mbc)$}
        \begin{fact}\label{id:lower_bound_mi}
            For any distribution $r(\mba|\mbb)$ and $a,b\sim p(\mba,\mathbf{b})$,
            \[H(\mba|\mbb) = -\E_{\mba,\mbb} \log r(\mba|\mbb) - \KL{p(\mba|\mbb)}{r(\mba|\mbb)}\]
            and therefore,
            \[H(\mba|\mbb) \leq -\E_{\mba,\mbb} \log r(\mba|\mbb)\]
            and equality holds when $r(\mba|\mbb) = p(\mba|\mbb)$
        \end{fact}
        \noindent
        By a direct application of above identity, we have:
        \begin{align}
            I(\mby:\mbz)&= H(\mby)-H(\mby|\mbz) \nonumber \\
                        &\geq H(\mby) +\max_r \E_{\mby, \mbz} \log r(\mby|\mbz)\label{eq:lower_bound_classifier}
            \intertext{and similarly,}
            I(\mby:\mbz|\mbc) &\geq H(\mby|\mbc) +\max_r \E_{\mby,\mbz,\mbc} \log r(\mby|\mbz,\mbc)
                \label{eq:lower_bound_conditional_classifier}
        \end{align}
        \noindent
        $H(\mby)$ and $H(\mby|\mbc)$ are properties of data and, therefore, constant from the optimization perspective. When $\mby$ is a one-dimensional variable denoting the target class, this is equivalent to minimizing cross-entropy. To this end, we will parametrize the variational distribution $r$ using a neural network with parameters $\psi$, but other models can also be used.

    \subsection{Upper Bound for $I(\mbz:\mbc)$}
        Our technique for upper-bounding $I(\mbz:\mbc)$ is similar to \citet{moyer2018invariant} and makes use of the following observations:
        \begin{equation}
            I(\mbz:\mbc) = I(\mbz:\mbc|\mbx) + I(\mbz:\mbx) - I(\mbz:\mbx|\mbc)
            \label{eq:mi_diff}
        \end{equation}
        $I(\mbz:\mbc|\mbx) = 0$, as $z$ is a function of $x$ and some independent noise. %
        As a result, we have  $I(\mbz:\mbc) = I(\mbz:\mbx) - I(\mbx:\mbz|\mbc)$. The first term is the information bottleneck term~\cite{alemi2016deep}
        and limits  the information about $\mbx$ in $\mbz$, and we will bound it by specifying a prior over $\mbz$. Second term tries to preserve information about $\mbx$ but not in $\mbc$ and we will lower bound it via contrastive estimation%
            .

        \subsubsection{Upper Bound for $I(\mbz:\mbx)$ by Specifying a Prior:}
            In order to upper-bound $I(\mbz:\mbx)$, we use the following observation:
            \begin{fact}\label{id:rate}
                For any $a,b\sim p(\mba,\mbb)$ and distribution $q(\mba)$%
                \[I(\mba;\mbb) = \E_{\mba,\mbb} \log \frac {p(\mba|\mbb)}{q(\mba)} - \KL{p(\mba)}{q(\mba)}\]
                and therefore,
                \begin{equation}
                    I(\mba:\mbb) \leq \E_{\mba,\mbb} \log \frac {p(\mba|\mbb)}{q(\mba)} = \E_\mbb \KL {p(\mba|\mbb)}{q(\mba)}
                    \label{eq:prior_upper_bound}
                \end{equation}
                and equality holds when $p(\mba) = q(\mba)$.
            \end{fact}\noindent
            Therefore,  we have
            \begin{equation}\label{eq:rate}
                I(\mbz:\mbx) \leq  \E_{\mbx} \KL {q(\mbz|\mbx;\phi)}{p(\mbz)}
            \end{equation}
            where $p(\mbz)$ is any distribution. This is similar to the rate term in the VAE or information bottleneck approach~\cite{alemi2016deep,Higgins}.
            Motivated by this similitude, we let $p(\mbz)$ be standard normal distribution and
            $q(\mbz|\mbx;\phi)$ be a diagonal gaussian distribution whose mean and variance are parametrized as $\mu(x)= f_\mu(x;\phi), \Sigma(x)= f_\Sigma (x;\phi)$. Other parameterizations like normalizing flows~\cite{rezende2015variational} or echo noise~\cite{brekelmans2019exact} can
            be used too.

        \subsubsection{Lower Bound for $I(\mbx:\mbz|\mbc)$ via Contrastive Estimation:}
            We propose to lower bound $I(\mbx:\mbz|\mbc)$ via contrastive estimation of mutual information, and we use the following proposition to derive our estimator.
            \begin{proposition} \cite{poole2019variational}
                \label{prop:conditional_nce}
                For any
                $u, v, w\sim p(\mbu, \mbv, \mbw)$,
                $\tilde u \sim p(\mbu|\mbw)$,
                and function $f$, we have
                \begin{equation}
                    I(\mbu: \mbv|\mbw) \geq \mathbb \E_{\mbu,\mbv,\mbw}\log  \frac {e^{f(u,v,w)}} {\frac 1 M \sum_{j=1}^M e^{f(\tilde u_j, v, w)}}\label{eq:contrastive_estimation}
                \end{equation}
                where,
                $u, \tilde u \in \mathcal U$,
                $v \in \mathcal V$,
                $w \in \mathcal W$,
                $f \mathrel{:}\mathcal U \times \mathcal V\times \mathcal W \rightarrow \mathbb R$
                and M is the number of samples from $p(\mbu|\mbw)$.
            \end{proposition}
            \noindent
            The proof is similar to the non-conditional version~\cite{poole2019variational}, but for completeness, we present the proof in the \full{Appendix~\ref{subsec:cond_nce_proof}}{Appendix A.2}. As a direct application of this result, we can lower-bound and maximize $I(\mbz:\mbx|\mbc)$. However, there is a caveat that we need to sample from $P(\mbz|\mbc)$. Sampling from this conditional distribution in the general case can be hard, but it can be easily accomplished for our problem.  Often for fairness applications, $\mbc$ is a discrete random variable with low cardinality; in fact, it is often a binary random variable. Therefore, we can consider $\{z_j \mathrel{:} (z_j, c_j=i) \} $  to be the  samples from $p(\mbz|\mbc=i)$. In our experiments, we parametrize $f(z, x, c)$ as a bilinear function (similar to \citet{cpc}).

            \subsection{Overall Variational Objective}
            Using Eq.~\ref{eq:mi_diff}, we can write our objective, i.e., Eq.~\ref{eq:conditional_objective} in terms of three information theoretic quantities as:
            \[
                \max_q
                    \underbrace{I(\mby:\mbz|\mbc)}_{\text{Eq.~\ref{eq:lower_bound_conditional_classifier}}}
                    - \beta \Big (
                        \underbrace{I(\mbz:\mbx)}_\text{Eq.~\ref{eq:rate}}
                        - \underbrace{I(\mbz:\mbx|\mbc)}_{\text{Eq.~\ref  {eq:contrastive_estimation}}}
                    \Big )
            \]
            The first term is lower-bounded using Eq.~\ref{eq:lower_bound_conditional_classifier} and it is the same as the  cross-entropy loss. Second term is upper bounded using Eq.~\ref{eq:rate}. For this, we set $p(\mbz)$ to be standard normal distribution and $q(\mbz|\mbx)$ is a normal distribution whose parameters are generated by a neural network. The KL divergence term in Eq.~\ref{eq:rate} can then be expressed in the closed form.     Finally, the last term is lower-bounded using the contrastive conditional mutual information estimator from Proposition~\ref{prop:conditional_nce} (Eq.~\ref{eq:contrastive_estimation}). We call our objective \textbf{F}air \textbf{C}ontrastive \textbf{R}epresentation \textbf{L}earner (FCRL).

            Objective~\ref{eq:objective} can be similarly computed and optimized using Eqs.~\ref{eq:lower_bound_classifier}, \ref{eq:rate}, and~\ref{eq:contrastive_estimation}. We provide comprehensive details to compute  the proposed variational objective in \full{Appendix~\ref{subsec:loss_implementation}}{Appendix C.1}.

            Even though objective~\ref{eq:objective} \&~\ref{eq:conditional_objective} require a single parameter $\beta$ to be varied, we found in our experiments that it is necessary to use different multipliers for $I(\mbx:\mbz|\mbc)$ and $I(\mbz:\mbx)$  to effectively tradeoff fairness and accuracy. Therefore, we introduce another parameter $\lambda$. The $I(\mbz:\mbc)$ term in both the objectives then becomes $I(\mbz:\mbx)- \lambda I(\mbx:\mbz|\mbc)$. We found that $\lambda=2$ performs better empirically and so we set $\lambda=2$, unless specified otherwise.

        \subsection{Other Bounds for $I(\mbz:\mbc)$ and Their Caveats}
            One may observe that $I(\mbz:\mbc)$ could also be upper-bounded as a direct consequence of Eq.~\ref{eq:rate}. This is possible, however upper bounding $I(\mbz:\mbc)$ simply by introducing a prior marginal $p(\mbz)$ as in Eq.~\ref{eq:prior_upper_bound} is not tractable, as that would require computing $p(\mbz|\mbc)$~\cite{song2019learning}. Therefore, it is necessary to decompose $I(\mbz:\mbc)$ as the difference of two information-theoretic quantities (Eq.~\ref{eq:mi_diff}).

        Another way to bound $I(\mbx:\mbz|\mbc)$ is to use the lower bound from Eq.~\ref{eq:lower_bound_conditional_classifier} which leads to a conditional reconstruction term.
        \begin{equation}\label{eq:cond_distortion}
            I(\mbx:\mbz|\mbc)\geq H(\mbx|\mbc)+ \max_\theta \E_{\mbx,\mbz,\mbc} \log q(\mbx|\mbz,\mbc;\theta)
        \end{equation}
        where  $q(\mbx|\mbz, \mbc;\theta)$ is some distribution parametrized by $\theta$.
        $H(\mbx|\mbc)$ is a constant. The last term can be seen as reconstructing $x$ from  $c$ and its latent representation $z$
        and is similar to the distortion term in VAE, albeit with conditioning on $\mbc$~\cite{moyer2018invariant}. While this is a tractable bound, it involves training a conditional decoder, which may be hard for certain domains and almost always requires either complex models or the restrictive assumption that feature dimensions are independent given $\mbz$. It also requires explicitly stating the distribution of $\mbx$, which is bypassed using contrastive estimation.  In our experiments, we found that this approximation is limiting (Fig.~\ref{fig:contrastive_vs_reconstruction}); therefore, we propose using a decoder-free variational approximation via contrastive estimation of mutual information.

        \subsubsection[]{Remarks on Adversarial Approach:}
        One may also use Eq.~\ref{eq:lower_bound_classifier} to approximate $I(\mbz:\mbc)$, which leads to a common technique to bound $I(\mbz:\mbc)$ via adversarial learning.  This is a lower bound to $I(\mbz:\mbc)$, but since we want to minimize $I(\mbz:\mbc)$, it should ideally be upper bounded. However, Eq.~\ref{eq:lower_bound_classifier} is often used and leads to adversarial min-max approaches~\cite{madras2018learning,edwards2015censoring,song2019learning}, where maximization is over parameters of the classifier $r(\mbc|\mbz)$ and minimization is over $q(\mbz|\mbx)$. Other than the difficulty of optimization, we can see that adversarial approaches have a fundamental problem: they minimize a \emph{lower bound} at each iteration. The gap between mutual information and its lower bound can be arbitrarily bad if the set of adversarial classifiers is restricted, as it must be in practice.
        Moreover, rather than using exact maximization, most of the methods approximate it using one or a few steps of SGD. This also puts into question one of the common approaches for evaluating invariant representations by predicting $c$ from $z$. Indeed, in our experiments, we find that this evaluation can be misleading and may pass unfair representations as fair.

\section{Experiments}\label{sec:experiments}
    \newcommand{\N}{5 }
    \begin{figure*}
        \centering
        \begin{subfigure}{0.49\textwidth}
            \centering
            \includegraphics[width=0.8\textwidth]{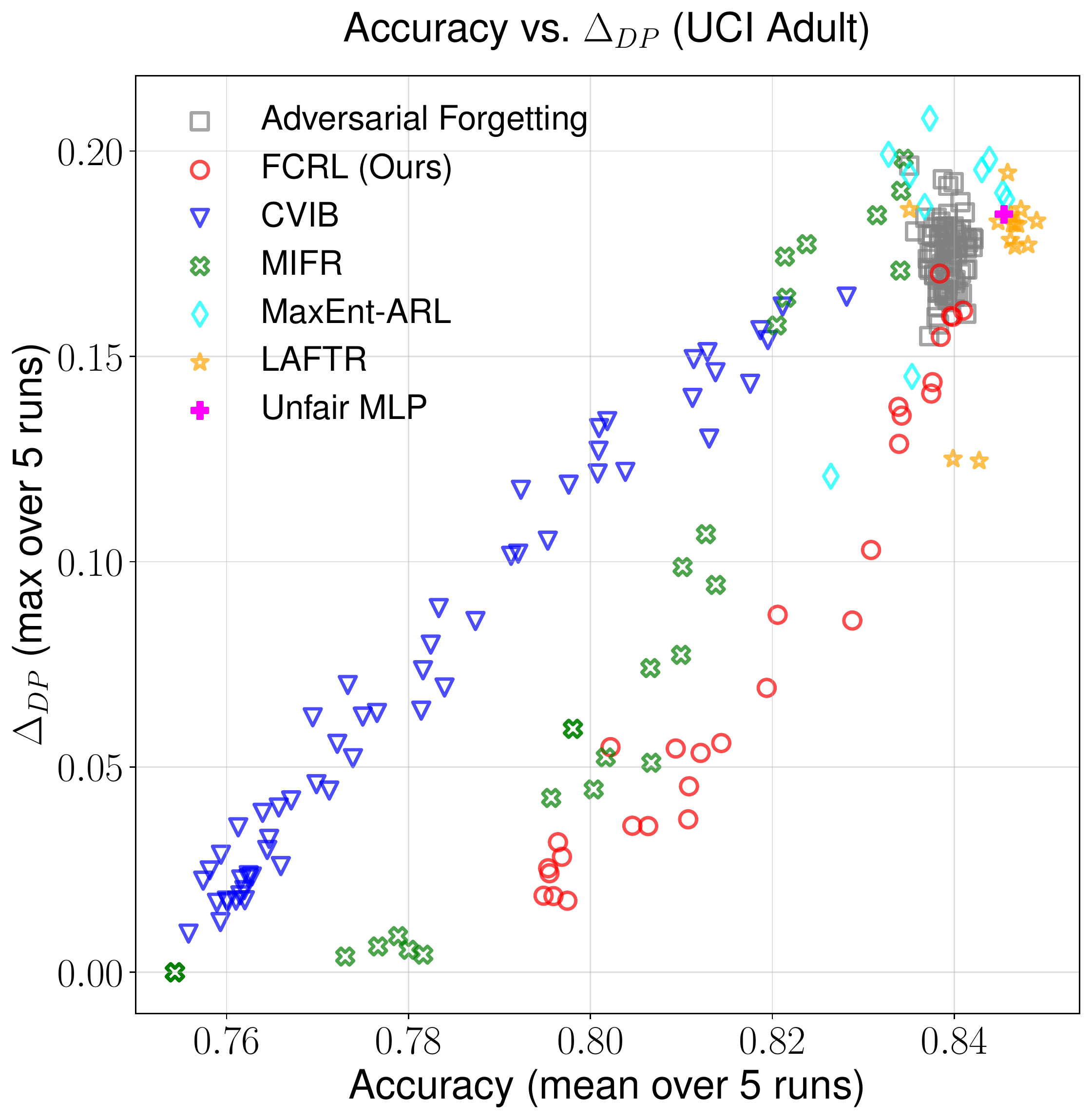}
        \end{subfigure}
        \begin{subfigure}{0.49\textwidth}
            \centering
            \includegraphics[width=0.8\textwidth]{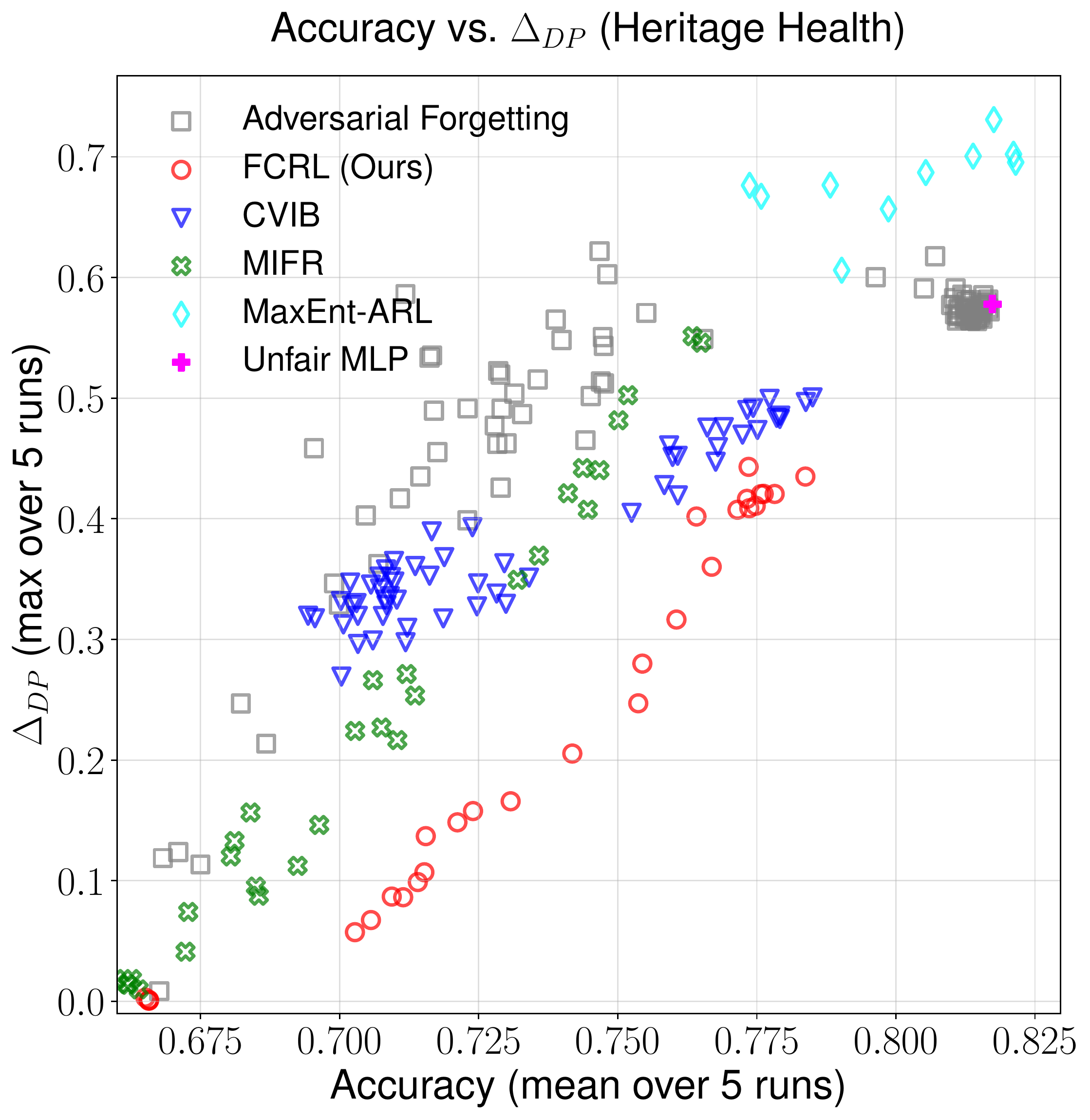}
        \end{subfigure}
        \caption{Parity vs.\ Accuracy trade-off for \textit{UCI Adult} and \textit{Heritage Health}  dataset using a 1-hidden-layer MLP. Lower $\Delta_{DP}$ is better, and higher accuracy is better. We use representations generated by varying each method’s inherent loss hyperparameters  to get different  points.
        See Table~\ref{tab:AOPAC_nn_1_layer} for quantitative results.
        Results with other decision algorithms are shown in \full{Appendix~\ref{sec:other_eval}}{Appendix D}.}\label{fig:baselines}
    \end{figure*}

    We validate our approach on two datasets --- \textit{UCI Adult }~\cite{Dua:2019} and \textit{Heritage Health}\footnote{\url{https://www.kaggle.com/c/hhp}} Dataset. \textit{UCI Adult} is 1994 census data with 30K samples in the train set and 15K samples in the test set. The target task is to predict whether the income exceeds \$50K, and the protected attribute is considered gender (which is binary in this case). We use the same preprocessing as \citet{moyer2018invariant}.  \textit{Heritage Health} dataset is data of around 51K patients (40K in the train set and 11K in the test set), and the task is to predict the Charleson Index, which is an indicator of 10-year survival of a patient. We consider age as the protected attribute, which has 9 possible values. We use the same preprocessing as \citet{song2019learning}.

    A fair representation learning algorithm aims to produce representations such that any downstream decision algorithm that uses these representations will produce fairer results.
    Therefore, similar to \citet{madras2018learning}, we train the representation learning algorithm on training data and evaluate the representations by training classifiers for downstream prediction tasks. Since our purpose is to assess the representations, we report average accuracy (as an indicator of most likely performance) and maximum parity (as an indicator of worst-case bias) computed over 5 runs of the decision algorithm with random seeds.
    Unlike \citet{madras2018learning}, we also allow for preprocessing to be done on representations. Preprocessing steps like min-max or standard scaling are common and often precede training of classifiers in a regular machine learning pipeline. The held-out test set is  used to evaluate representations on downstream tasks only; we use the training set for all the other steps.

    We compare with a number of recent approaches, including information-theoretic and adversarial methods from the recent literature. MIFR~\cite{song2019learning}  is a recent approach demonstrating competitive performance. MIFR combines information-theoretic and adversarial methods and generalizes several previous fair representation learning approaches~\cite{Louizos2015,edwards2015censoring,madras2018learning,Zemel2013}. A related approach, CVIB~\cite{moyer2018invariant}, is based solely on information-theoretic optimization without adversarial training. We also compare with recent state-of-the-art adversarial methods. In particular, we compare with Adversarial Forgetting~\cite{jaiswal2020invariant}, which is the state-of-the-art adversarial method for learning invariant representations. \citet{roy2019mitigating} (MaxEnt-ARL) proposes a theoretically superior min-max objective by proposing to  train the encoder  to maximize the entropy of sensitive attributes. In contrast, traditional adversarial approaches try to minimize the likelihood of the discriminator.
    Finally, we compare with LAFTR~\cite{madras2018learning}, which minimizes a more directed adversarial approximation to parity and has been designed bearing in mind parity-accuracy trade-offs. However, LAFTR is only applicable when $\mbc$ is a binary variable.  As a baseline, we also train a one hidden layer MLP predictor directly on the data without regards to fairness (\emph{Unfair MLP}).

    We visualize the trade-offs between fairness and task performance by plotting parity vs.\ accuracy curves for each representation learning algorithm.
    We vary each method's inherent hyperparameters over the ranges specified in the original works  to get different points on this curve.
    For a fair comparison, we set $d\equal 8$ for all the methods
    and use model components like encoder, decoder, etc.,\ of the same complexity.
    We provide a detailed list of the hyperparameter grids, network architectures, and other training setups for all the methods in  \full{Appendix~\ref{sec:training_details}}{Appendix D}. The code to reproduce all the experiments is available at \codeurl.
    We give a quantitative summary of performance across the entire spectrum of trade-offs by reporting the area over the Parity-Accuracy curve. This section uses a 1-hidden-layer MLP with ReLU non-linearity and 50 neurons in the hidden layer as the decision algorithm and representations are preprocessed by standard scaling.
    Results with other decision algorithms, i.e., Random Forest, SVM, 2-hidden-layer MLP, and logistic regression, are shown in the  \full{Appendix~\ref{sec:other_eval}}{Appendix D}.

        \subsubsection{Improved Accuracy Versus Parity Trade-offs:}
            For different fair representation learners, we compare accuracy versus parity achieved for a specific downstream classifier. The goal is to push the frontier of achievable trade-offs as far to the bottom-right as possible, i.e., to achieve the best possible accuracy while maintaining a low parity.
            From a visual inspection of Fig.~\ref{fig:baselines}, we can see that our approach preserves more information about label $\mby$, across a range of fairness thresholds for both \textit{UCI Adult} and \textit{Heritage Health} datasets.
            We observed improved trade-offs for our method even when different downstream classifiers are used (results are shown in  \full{Appendix~\ref{sec:other_eval}}{Appendix D}), matching expectations since our method gives theoretical bounds on parity that hold regardless of the classifier used. In contrast, methods that use adversaries in training see increases in parity when a more powerful classifier is used for downstream classification.

        \subsubsection{Metric Quantifying Performance Across Trade-offs:}
            \begin{table}[b]
	\centering
	\begin{tabular}{lcc}
		\toprule
		Method & UCI Adult & Heritage Health \\
		\cmidrule(r){1-1} \cmidrule(lr){2-2} \cmidrule(lr){3-3}
		FCRL (Ours) & \textbf{0.307} & \textbf{0.319} \\
		CVIB & 0.182 & 0.191 \\
		MIFR & 0.251 & 0.202 \\
		MaxEnt-ARL & 0.144 & 0 \\
		LAFTR & 0.155 & N/A \\
		Adversarial Forgetting & 0.087 & 0.156 \\
		\bottomrule
	\end{tabular}
	\caption{Area Over Parity-Accuracy Curve (Higher is better)}
	\label{tab:AOPAC_nn_1_layer}
\end{table}

            Fair representation learning approaches often provide a single number to quantify fairness by predicting $c$ from $z$ at a single accuracy level or only qualitatively demonstrate results by showing the parity-accuracy curve.  We report the normalized area over the parity-accuracy curve in Table~\ref{tab:AOPAC_nn_1_layer} (1 being the maximum value) to enable quantitative comparison.
            It is the normalized hypervolume of the feasible parity-accuracy region, and an efficient fair representation learning method should maximize it.
            We provide more intuition and details to compute this metric in   \full{Appendix~\ref{sec:aopac}}{Appendix F}.

        \subsubsection{Controlling Parity:}
\begin{figure}[t!]
                \centering
                \includegraphics[width=0.39\textwidth]{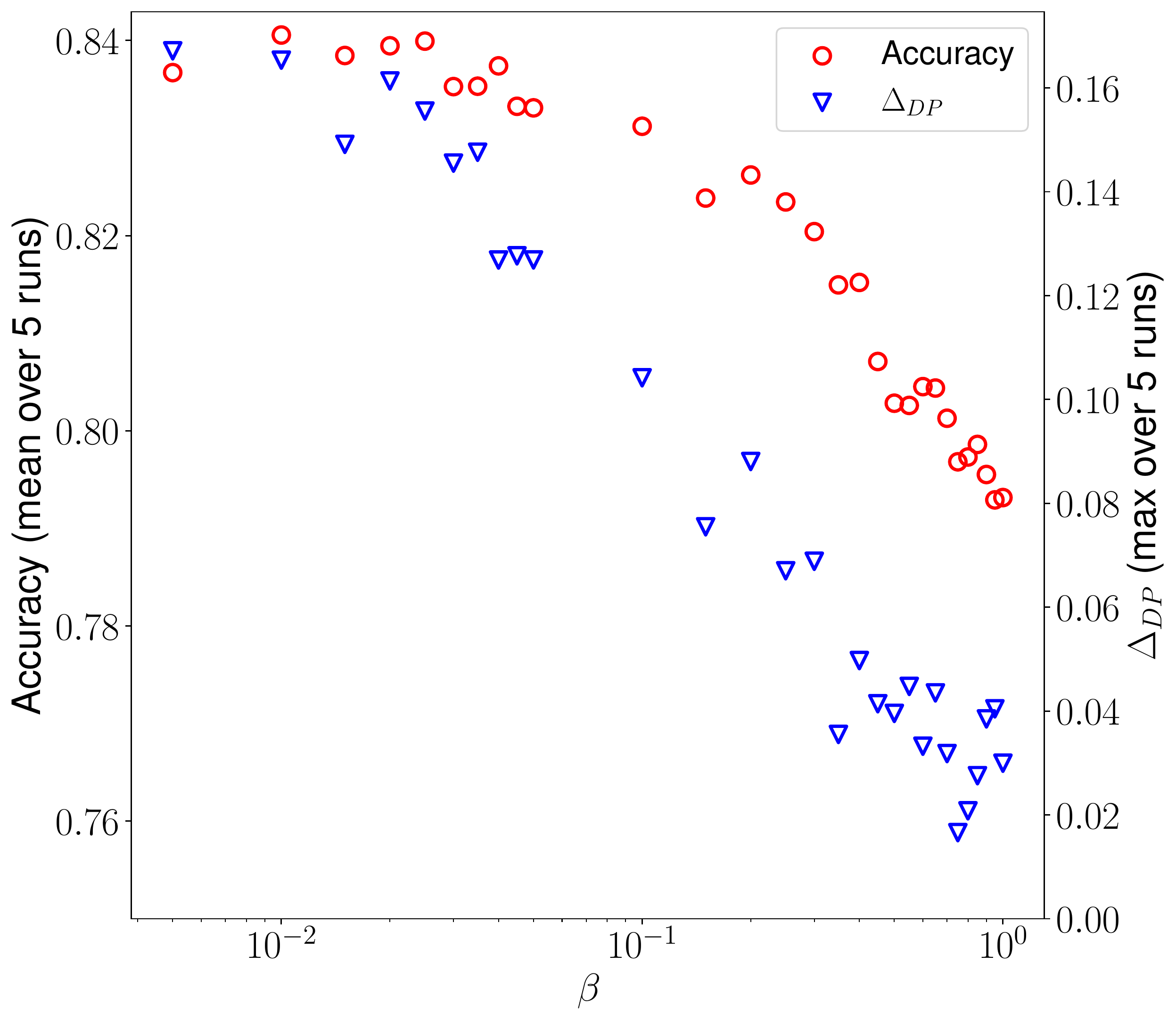}
                \caption{Parity and Accuracy variation with $\beta$ for  \textit{UCI Adult} dataset. Our method can explore feasible regions of parity and accuracy by varying only a single parameter $\beta$. }\label{fig:beta_curve}
            \end{figure}

            Our approach has a single intuitive hyperparameter $\beta$, which can be used to  control $I(\mbz :\mbc)$  directly and, therefore, via Thm.~\ref{thm:parity_mi_relation}, to monotonically control parity (see Fig.~\ref{fig:beta_curve}). For the \textit{UCI Adult} dataset, MIFR~\cite{song2019learning} is competitive with our approach near low parity; however,  it fails to achieve higher accuracy at higher demographic parity. This is because MIFR proposed to use $I(\mbz:\mbx)$ as an upper bound to $I(\mbz:\mbc)$, which is very loose, and this penalizes information about $x$ as well, which is not desirable. CVIB~\cite{moyer2018invariant} is able to consistently trade-off accuracy while using a reconstruction based bound. But it maximizes $I(\mby:\mbz)$ which conflicts with desired minimization of $I(\mbz:\mbc)$ (see Sec.~\ref{subsec:interference}).

        \subsubsection{Fine-tuning the Parity:}

            To achieve different points on parity-accuracy trade-off, one must train models with different loss hyperparameters ($\beta$ in our case). To a practitioner, parity-accuracy curves are essential, as it helps in deciding which representations should be used to satisfy the desired fairness constraints. The additional computational cost of training models, as well as the introduction of additional hyperparameters for fair learning  pose significant hindrances towards the adoption of fair models~\cite{sylvester2018appliedfairness}.

\begin{figure}
                \centering
                \includegraphics[width=0.35\textwidth]{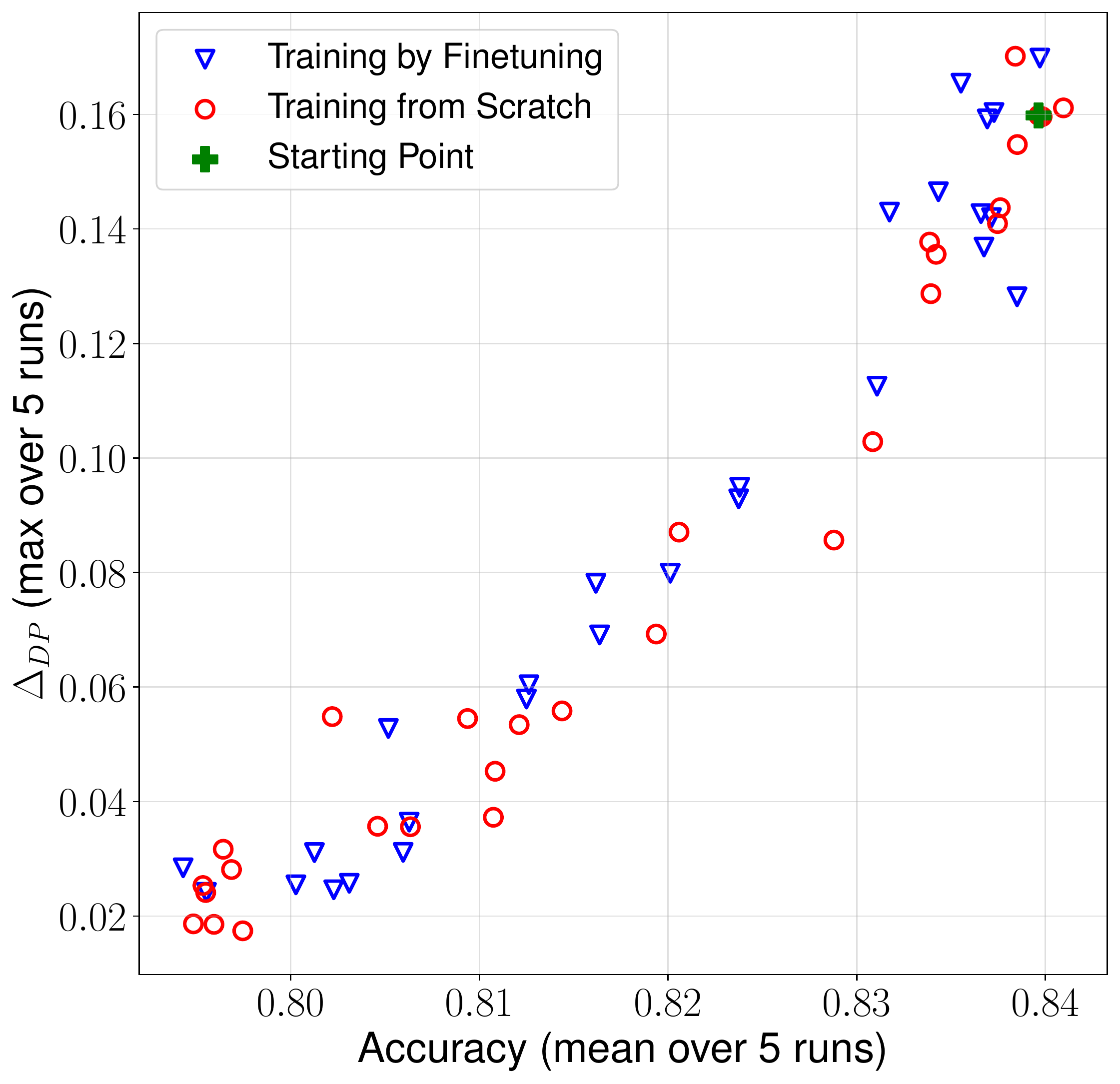}
                \caption{Exploring parity-accuracy trade-off for \textit{UCI Adult} dataset by varying $\beta$ and fine-tuning the model for 20 epochs. When training from scratch, we need to train for 6000 epochs  to generate 30 points (200 epochs each). However, with fine-tuning we can get the same result with only 780 epochs $(200 + 29\times 20)$.}\label{fig:finetuning}
            \end{figure}

            Recently, \citet{Gao2020} showed that it is possible to explore different equilibrium points on the rate-distortion curve with iso-accuracy constraints by adjusting the loss coefficients according to some dynamical process and finetuning the model parameters.
            Similarly, we show that we can finetune a trained model to produce different points on the parity-accuracy curve. Since our loss function has a single coefficient ($\beta$), we do not need to derive the dynamical process. We can simply vary $\beta$ by sufficiently small steps and explore the parity-accuracy trade-off by finetuning the trained model. We can see that our method performs equally well when finetuning compared to training from scratch (See Fig.~\ref{fig:finetuning}), reducing the computational cost drastically. We expect that the reduced computational cost of achieving the parity-accuracy curve and introduction of only a single intuitive hyperparameter along with improved parity-accuracy trade-off should reduce resistance towards the adoption of fair models.

        \subsubsection{Uncovering Hidden Bias:}
            In our experiments, we find that adversarial approaches that use deterministic representations (e.g., \citet{roy2019mitigating,jaiswal2020invariant}) are particularly susceptible to hidden bias, which can be revealed by slightly modified classifiers that consistently result in large parity values.
            \citet{roy2019mitigating} demonstrated that the representations produced were fair w.r.t to $c$ by showing that an MLP trained to predict $c$ from $z$ performs poorly. This evaluation approach of predicting $c$ from $z$ is often used to demonstrate fairness or invariance of the representation w.r.t to $\mbc$~\cite{moyer2018invariant,jaiswal2020invariant,xie2017controllable}. However, we emphasize that due to Eq.~\ref{eq:lower_bound_classifier}, this is only a lower bound on $I(\mbz:\mbc)$ and, therefore, any claims relying solely on this metric are weak. Further, we found
            that these methods do not remove information about sensitive attributes and, at best, obfuscate it such that an adversary cannot easily recover the protected attribute during training. We found that even if a classifier trained directly on $z$ cannot predict $c$, after preprocessing $z$ by standard scaling, it becomes very easy to predict the sensitive attribute even with a model of similar or lower complexity as the original
            adversary used during the training. As a result, some of these approaches could not achieve low parities and even exhibited higher parity than the baseline \emph{Unfair MLP} trained without fairness constraints (see Fig.~\ref{fig:baselines}).
            We investigate this further in the \full{Appendix~\ref{sec:hidden_bias}}{Appendix E}.

    \subsection*{Ablation Study of Information-Theoretic Methods}
    Next, we will show that our proposed objective of maximizing $I(\mby:\mbz | \mbc)$ is more efficient at exploring parity-accuracy trade-offs compared to maximizing $I(\mby:\mbz)$ (i.e.,\ objective~\ref{eq:objective} vs.~\ref{eq:conditional_objective}).  We also compare the effect of optimizing the reconstruction based information bound (Eq.~\ref{eq:cond_distortion}) with our proposed contrastive information estimator (Eq.~\ref{eq:contrastive_estimation}).

        \subsubsection{Maximizing $I(\mby:\mbz|\mbc)$ vs. $I(\mby:\mbz)$:}
            We see in Fig.~\ref{fig:cond_vs_non_cond} that when maximizing $I(\mby:\mbz)$, we can still reduce the parity, but this comes at the cost of predictive performance. When using $I(\mby:\mbz|\mbc)$, we see that the parity sharply drops with only a slight drop in the accuracy,  and we see much better trade-offs between predictability and fairness. By conditioning mutual information on $c$, we see that $z$ can easily retain information about $y$ without conflicting with $I(\mbz:\mbc)$ objective as explained in Sec.~\ref{subsec:interference}.

        \subsubsection{Reconstruction vs. Contrastive Estimation:}
\begin{figure*}[bt!]
                \centering
                \begin{subfigure}[t]{0.49\textwidth}
                    \centering
                    \includegraphics[width=0.74\textwidth]{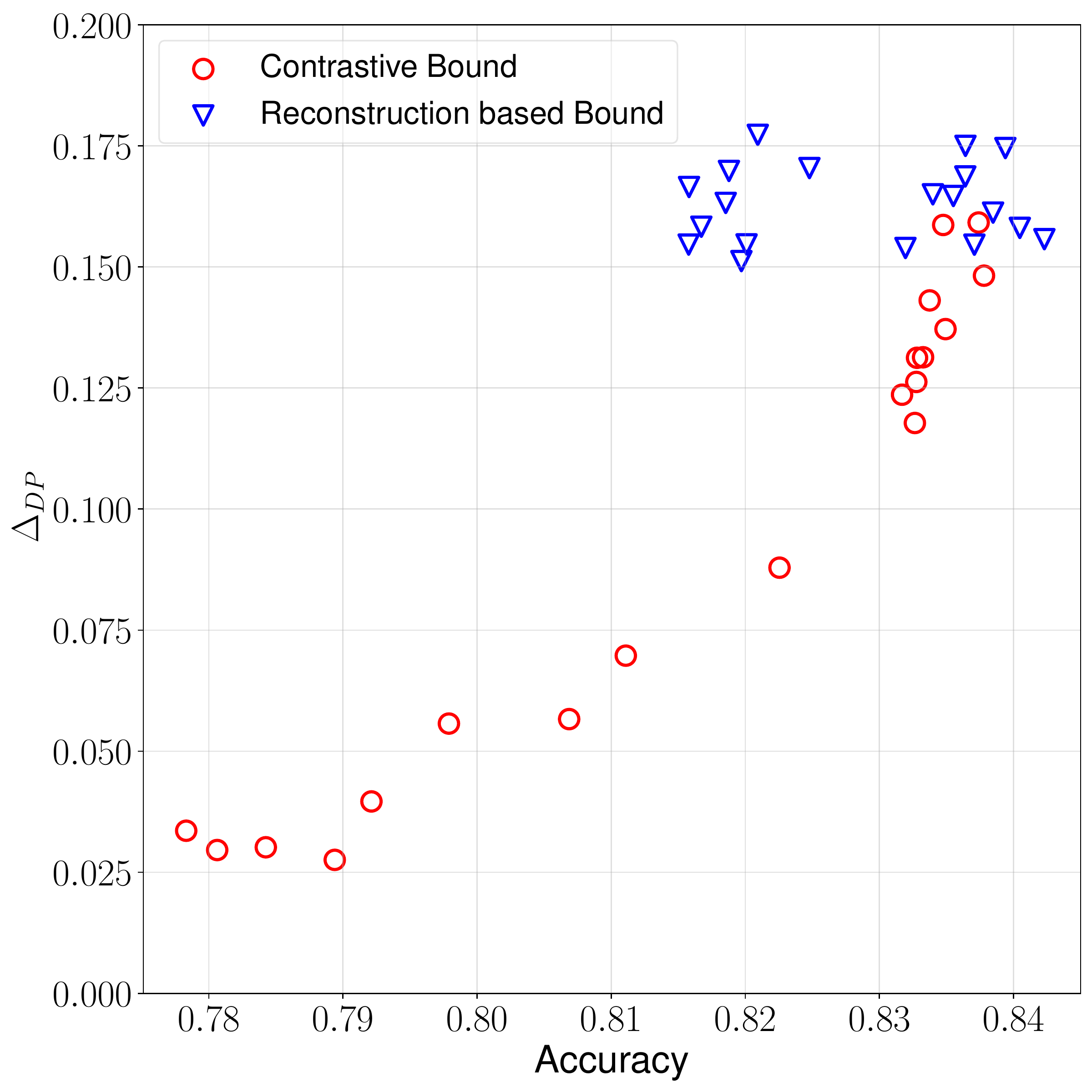}
                    \subcaption{} \label{fig:contrastive_vs_reconstruction}
                \end{subfigure}
                \begin{subfigure}[t]{0.49\textwidth}
                    \centering
                    \includegraphics[width=0.74\textwidth]{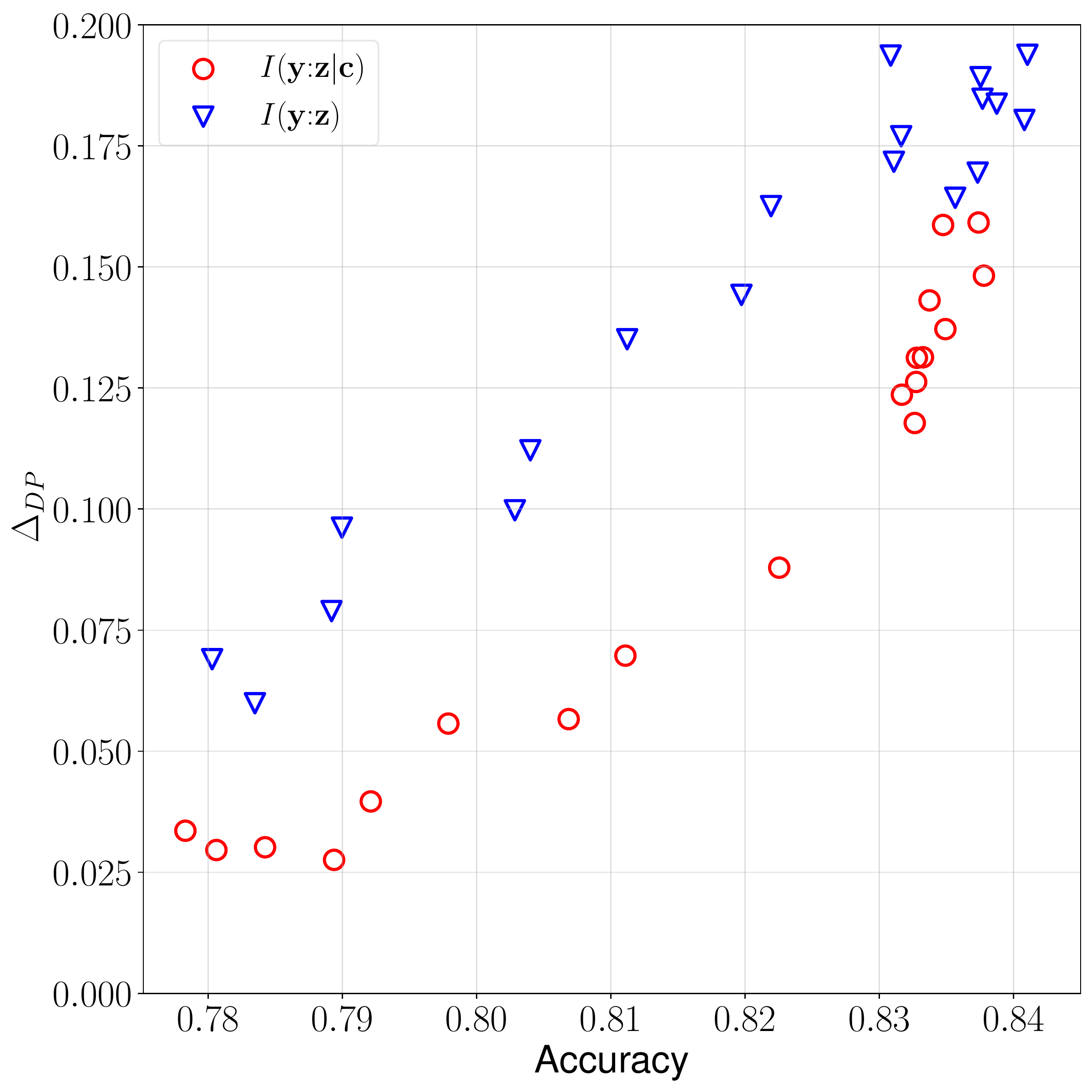}
                    \subcaption{}\label{fig:cond_vs_non_cond}
                \end{subfigure}
                \caption{Ablation studies on \textit{UCI Adult} dataset: Reconstruction (Eq.~\ref{eq:cond_distortion}) and Contrastive (Eq.~\ref{eq:contrastive_estimation}) mutual information estimation bounds are compared in Fig.\ a. Fig.\ b compares Objective~\ref{eq:conditional_objective} and Objective~\ref{eq:objective}. $\lambda$ was set to 1 for these experiments.}

            \end{figure*}

            Fig.~\ref{fig:contrastive_vs_reconstruction} compares optimizing reconstruction and contrastive estimation based bounds for $I(\mbx:\mbz|\mbc)$. Reconstruction based bounds on information rely on explicitly modeling the data distribution through a decoder, $q(\mbx|\mbz, \mbc)$, that reconstructs the data. For these methods to be effective, they must give reasonable reconstructions and therefore preserve most of the information in the data.  We see in the results that preserving information leads to high accuracy but also high parity. Contrastive estimation is better at achieving lower parity
            than reconstruction based bounds because it can directly model high-level features that are predictive and fair without requiring a model to reconstruct data.

\section{Related Work}\label{sec:related}
As machine learning becomes more enmeshed in everyday life, a vibrant field of fair machine learning research has emerged to question potential risks~\cite{Mehrabi2019}. Different applications may prescribe different notions of fairness, and these notions may even conflict with each other~\cite{Kleinberg2016inherent}. In this work, we focus on fairness between groups concerning decisions that may be cast as classification problems. In particular, our work is focused on controlling statistical parity or disparate impact, which was proposed in~\citet{Dwork2011} and is widely adopted in the literature~\cite{edwards2015censoring, McNamara}. Statistical parity is preferred as a fairness measure  when the goal is to ensure equal selections from different groups. One such example is the US government's \emph{Uniform Guidelines on Employee Selection Procedure}~\cite{USEOEC} adoption of 80\% rule to ensure fairness in selection procedures. Statistical parity as a fairness measure also captures a lay person's perception of fairness~\cite{srivastava2019}. Incorporating other popular notions of group fairness like Equalized-Opportunity and Equalized-Odds~\cite{Hardt2016}, and notions of individual fairness (e.g.,~\citet{sharifi2019_aif}) in our approach is left as future work.

Fair classification methods are often categorized based on which stage of the machine learning pipeline they target.
Post-processing methods work by modifying the
trained classifier (e.g.,~\citet{Hardt2016}), in-processing methods work by regularizing the classifier during training to ensure fairness (e.g.,~\citet{zafar2017fairness}), and pre-processing methods transform the original dataset to reduce bias present in the data. Pre-processing methods are beneficial when the onus of fairness is on a third party or the data controller, and the end-user may be oblivious to fairness constraints~\cite{McNamara, cissefairness}. Our work also belongs to this category, and we specifically focus on controlling parity.

While post-processing and in-processing methods have to ensure fairness for a specific classifier, pre-processing methods must ensure fairness with respect to any downstream classification algorithm. Many pre-processing methods have discussed the desiderata of ensuring strong guarantees on fairness so that any downstream classifier may be used freely~\cite{McNamara,song2019learning,edwards2015censoring,madras2018learning}. However, their operationalization often leads to an approach that may not ensure guarantees (due to limits of adversarial methods, for instance). Our approach to these problems bounds parity
in terms of mutual information, and then we optimize tight upper bounds on this information. Other works have explored information-theoretic objectives for learning invariant representations~\cite{moyer2018invariant} and fair representations~\cite{song2019learning}. \citet{Dutta} use tools from information theory to analyze the trade-off between fairness and accuracy.
\citet{yuziGeometric} minimize correlation between the sensitive attributes and the representations to achieve fairness. Correlation is a linear measure of dependence. Therefore, their method provides no guarantees when the downstream classifier is non-linear. In contrast, our approach controls mutual information between the sensitive attribute and the representation, a more robust measure of dependence.

Contrastive learning and its variants have shown promising results for learning representations for many applications, e.g., images and speech~\cite{cpc}, text~\cite{mikolov2013distributed} and graphs~\cite{dgim}. We are the first to explore its application for learning fair representations.
Contrastive learning has been most actively explored in self-supervised learning, where the information to optimize is chosen by hand to be similar to some target task~\cite{Chen2020, cpc}, while in our work,
we demonstrated a natural connection between parity and mutual information.
Other variational bounds on information~\cite{poole2019variational} and estimators like MINE~\cite{belghazi2018mine} and NWJ~\cite{nguyen2010estimating} could also be leveraged for parity control using our results.

\section{Conclusion}
As the negative repercussions of biased data become increasingly apparent, governments and other organizations will require tools capable of controlling this bias. In many cases, compromises between fairness and task objectives will be desired and sometimes even legally enshrined in terms of required bounds on fairness measures like statistical parity. Many methods do not provide a way to control parity, and even if they do, often, it is  only in a heuristic way.  Adversarial classifiers are popular for checking bounds on parity, but these methods are not guaranteed against the possibility of a more powerful classifier either intentionally or accidentally exploiting bias still hidden in the data. By proving a one-to-one relationship between information-theoretic quantities and statistical parity of arbitrary classifiers, we can finally see how varying a single hyper-parameter controlling information can explore the entire fairness versus accuracy spectrum. This information-theoretic characterization is algorithm-independent so that our control of parity can be guaranteed regardless of downstream applications.

\section*{Acknowledgements}
We thank anonymous reviewers, especially reviewer 3 and 6 for their constructive feedback towards improving this paper.
This material is based upon work supported by the Defense Advanced Research Projects Agency (DARPA) under Agreement No.
\texttt{HR00112090104} and the DARPA NGS2 research award \texttt{W911NF-16-1-0575}.
AF \& BD were supported by NSF grant \#\texttt{1935451}. We would like to thank Shobhit Jain for spotting an error in our Thm.~\ref{thm:parity_mi_relation}.
\section{Ethical \& Broader Social Impact}

Machine learning is increasingly ubiquitous in daily life and can affect small things like what websites and products an individual sees to big things like who is approved for a loan or released on bail.
Because machine learning relies on training data that may reflect historical bias, the outcomes of small and large decisions based on these systems may propagate that bias.
The sudden emergence of a large amount of research on fair learning reflects the urgency and complexity of the issues. While clear-cut solutions to age-old ethical conundrums are unlikely, this should not prevent us from identifying spheres of human activity where machine-assisted decision-making can be made more fair.
Our work follows in this vein by looking for ways to improve parity when decisions affect two groups, while maintaining as much ability to predict good outcomes as possible.

Fair machine learning research, including this paper, is not without pitfalls. There is a danger that mathematically attractive principles for fairness can lead to a sense of false security. Depending on the context, these measures can lead to counter-intuitive results. It is difficult for humans to agree on a fair outcome in many situations, and we cannot expect a single formula to settle the debate.
For instance, \citet{Hardt2016} make two arguments against parity, the measure investigated in our paper. First, they argue that parity does not ensure intuitive notions of fairness because one way it can be satisfied is by taking qualified candidates from group A and an equal number of random candidates from group B. Even though parity is achieved, this may seem unfair as the most qualified candidate from group B could be omitted. Second, they argue that perfect parity hobbles the utility of machine learning. Because of subtle correlations among protected attributes and other predictive attributes, we end up completely destroying our ability to predict useful outcomes.

While the criticisms of parity given by \citet{Hardt2016} are valid, we believe that our work makes substantial contributions toward addressing both criticisms. First, we showed that trade-offs between parity and predictive performance could be significantly improved. This is important for wide-spread adoption of fair learning techniques, as voluntary adoption may be facilitated if the cost to performance can be demonstrated to be very low. Second, we introduced a tight connection between an information-theoretic perspective on fairness by limiting mutual information and parity. The generality of mutual information precludes bad faith solutions to achieve perfect parity like choosing qualified people from one group and random people from another. The reason for this is that, after learning a representation with bounded mutual information, information theory guarantees that it is impossible for any method to distinguish which group an individual is from in a way that would allow unequal application of selection criteria.

Although we believe that the information-theoretic approach we have provided in this paper is a more effective approach for guaranteeing parity,
it should be recognized that parity is not necessarily the ``fair'' solution in every context. For instance, to address the negative repercussions that historical bias has had on a group may require treatment that purposely violates parity~\cite{mehrabiequity}.

\nocite{gnuparallel}
\bibliography{references.bib}
\iffull%
   \cleardoublepage
\onecolumn
\appendix
\noindent
\begin{center}
    {\Large \bf Supplementary: Controllable Guarantees for Fair Outcomes
    via\\ Contrastive Information Estimation}
\end{center}

\newcommand\numbereqn{\addtocounter{equation}{1}\tag{\theequation}}
\setlength{\parskip}{0.5\baselineskip}%

\section{Proofs}\label{sec:proofs}
\newcommand{\JSD}{\text{JSD}}
\subsection{Proof of Theorem~\ref{thm:parity_mi_relation}}\label{subsec:mi_parity_relation_proof}

    \begin{theorem}\label{thm:parity_mi_relation_restated}
        \textbf{(Theorem~\ref{thm:parity_mi_relation} restated)}
        For some $z, c\sim p(\mbz,\mbc)$, $z\in\real^d$, $c\in \{0,1\}$, and any decision algorithm $\mathcal A$ that acts on $z$, we have
        \[I(\mbz:\mbc)\geq g\left(\pi, \DP\left(\mathcal A,\mbc\right)\right)\] where $\pi=P(\mbc= 1)$ and   $g$ is a strictly increasing  non-negative convex function in $\DP\left(\mathcal A,\mbc\right)$.
    \end{theorem}


    \begin{proof}
        \newcommand{\der}{\operatorname{d\!}}

        First, we will show that parity of any algorithm $\mathcal A$ that acts on $\mbz$  is upper bounded by  the variational distance between conditional distributions, $p(\mbz\mid\mbc\equal 1)$ and $p(\mbz\mid\mbc\equal 0)$.
        \begin{align*}
            \intertext{We know for some $\mathcal A$ acting on $z$,}
            \DP(\mathcal A, \mbc)
                &= \left|
                    P(\hat \mby\equal 1\mid \mbc\equal 1)- P(\hat \mby\equal 1\mid\mbc\equal 0)
                \right|\\
                &= \left|
                    \int_{z}\der z P(\hat \mby\equal 1 \mid \mbz )p(\mbz|c\equal 1) -
                    \int_z \der z P(\hat \mby\equal 1 \mid \mbz)p(\mbz\mid \mbc\equal 0)
                \right|\\
                &= \left|
                    \int_{z}\der z P(\hat \mby\equal 1 \mid \mbz)
                    \left\{p(\mbz\mid\mbc\equal 1) - p(\mbz\mid\mbc\equal 0)\right\}
                \right|\\
                &= \left|
                    \int_{z}\der z  P(\hat \mby\equal 1 \mid \mbz)
                    \left\{p(\mbz\mid\mbc\equal 1) - p(\mbz\mid\mbc\equal 0)\right\}
                \right|\\
                &\leq
                    \int_{z} \der z P(\hat \mby\equal 1 \mid \mbz)
                    \left| p(\mbz\mid\mbc\equal 1) - p(\mbz\mid \mbc\equal 0)
                \right |\\
                &\leq \int_{z} \der z
                    \left| p(\mbz\mid\mbc\equal 1) - p(\mbz\mid\mbc\equal 0)
                \right |&\text{Since, $P(\hat \mby=1\mid \mbz)\leq 1$}\\
                &= V(p(\mbz\mid\mbc\equal 0), p(\mbz\mid\mbc\equal 1) ) \numbereqn \label{eq:variational_distance_and_parity}
        \end{align*}
        where,
        \[
            V(p(\mbz\mid\mbc\equal 0), p(\mbz\mid\mbc\equal 1))=
            \int_{z} \der z \left|
                p(\mbz\mid\mbc\equal 1) - p(\mbz\mid\mbc\equal 0)
            \right | = \|p(\mbz\mid\mbc\equal 1) - p(\mbz\mid\mbc\equal 0) \|
        \]
        is the variational distance between $p(\mbz\mid\mbc\equal 1)$ and $p(\mbz\mid\mbc\equal 0)$.
        Next, we will show that mutual information, $I(\mbz:\mbc)$ is lower bounded by a strictly increasing function of variational distance between $p(\mbz\mid\mbc\equal 1)$ and $p(\mbz\mid\mbc\equal 0)$, and  therefore, by transitivity, also lower bounded by the function of parity of any $\mathcal A$.

        \noindent
        Letting $\pi=P(\mbc=1)$, we can write:
        \begin{align*}
            I(\mbz:\mbc)
            &= \E_{\mbz,\mbc} \log \cfrac {p(\mbz, \mbc)}{p(\mbz)p (\mbc)}\\
            &= \E_{\mbz,\mbc} \log \cfrac {p(\mbz\mid \mbc)}{p(\mbz)}\\
            &= (1-\pi) \E_{\mbz\mid\mbc=0} \log \cfrac {p(\mbz\mid\mbc=0)}{p(\mbz)} + \pi \E_{\mbz\mid\mbc=1} \log \cfrac {p(\mbz\mid\mbc=1)}{p(\mbz)}\\
            &=(1-\pi) \KL{p(\mbz\mid \mbc=0)}{p(\mbz)}+\pi\KL {p(\mbz\mid \mbc=1)}{p(\mbz)} \label{eq:jsd_mi} \numbereqn\\
            &=\JSD_{(1-\pi, \pi)}\left(p(\mbz\mid \mbc=0), p(\mbz\mid \mbc=1)\right)
        \end{align*}
        Last step is due to
        \begin{equation}
            p(\mbz) = \sum_c p(\mbz,\mbc)
            = (1-\pi) p(\mbz|\mbc=0) + \pi p(\mbz|\mbc=1)
            \label{eq:pz}
        \end{equation}
        and here $\JSD_{(1-\pi, \pi)}(p_1, p_2)$ denotes generalized Jensen-Shannon divergence with mixture weights $(1-\pi,\pi)$~\cite{lin1991divergence}.

        \noindent
        We know from \citet{toussaint1975sharper} that,
        \begin{equation}
            \KL{p_1}{p_2} \geq \max\left (\log \left( \frac{2+V}{2-V}\right) - \frac{2V}{2+V},  \frac{V^2}{2\ }+\frac{V^4}{36}+\frac{V^6}{288}\right) = f(V) \label{eq:variational_kl_inequality}
        \end{equation}
        For simplicity, we have used $V$ to denote variational distance $V(p_1, p_2)$. $f$ is defined in range $[0, 2)$. We note two important properties of function $f$ that are useful for our proof.
        \begin{itemize}
            \item $f$ is a non-negative strictly increasing function.
            \item $f$ is maximum of two convex functions and therefore also convex.
        \end{itemize}
        \noindent
        Combining Eqs.~\ref{eq:jsd_mi} and~\ref{eq:variational_kl_inequality}, and noting that,
        \begin{align*}
            V(p(\mbz\mid\mbc\equal 0), p(\mbz)) &= \|p(\mbz\mid\mbc\equal 0)-p(\mbz)\| = \pi\|p(\mbz\mid\mbc\equal 0)-p(\mbz\mid\mbc\equal 1)\|\\
            V(p(\mbz\mid\mbc\equal 1), p(\mbz)) &= \|p(\mbz\mid\mbc\equal 1)-p(\mbz)\| =( 1-\pi)\|p(\mbz\mid\mbc\equal 0)-p(\mbz\mid\mbc\equal 1)\|
        \end{align*}
        we get the required result,
        \begin{align*}
            I(\mbz:\mbc)
                &\geq (1-\pi) f \left(V(p(\mbz\mid\mbc\equal 0), p(\mbz))\right)+
                \pi f \left(V(p(\mbz\mid\mbc\equal 1), p(\mbz))\right)\\
            I(\mbz:\mbc)
                &\geq
                    (1-\pi) f \left(\pi V(p(\mbz\mid\mbc\equal 0), p(\mbz\mid\mbc\equal 1))\right) +
                    \pi f \left((1-\pi) V(p(\mbz\mid\mbc\equal 0), p(\mbz\mid\mbc\equal 1))\right)\\
                & \geq(1-\pi) f \left(\pi\DP(\mathcal A, \mbc)\right) +
                \pi f \left((1-\pi) \DP(\mathcal A, \mbc)\right) \\
                \intertext{(by Eq.~\ref{eq:variational_distance_and_parity} and  strictly increasing nature of $f$)}
                & = g(\pi, \DP(\mathcal A, \mbc))
        \end{align*}
        $g$ is positive weighted combination of non-negative strictly increasing convex functions and therefore also strictly increasing, non-negative, and convex. This completes the proof. Next, we visualize above Thm. in Fig.~\ref{fig:mi_parity_bound}
    \end{proof}

    \begin{figure}[h!]
        \centering
        \includegraphics[width=0.5\textwidth]{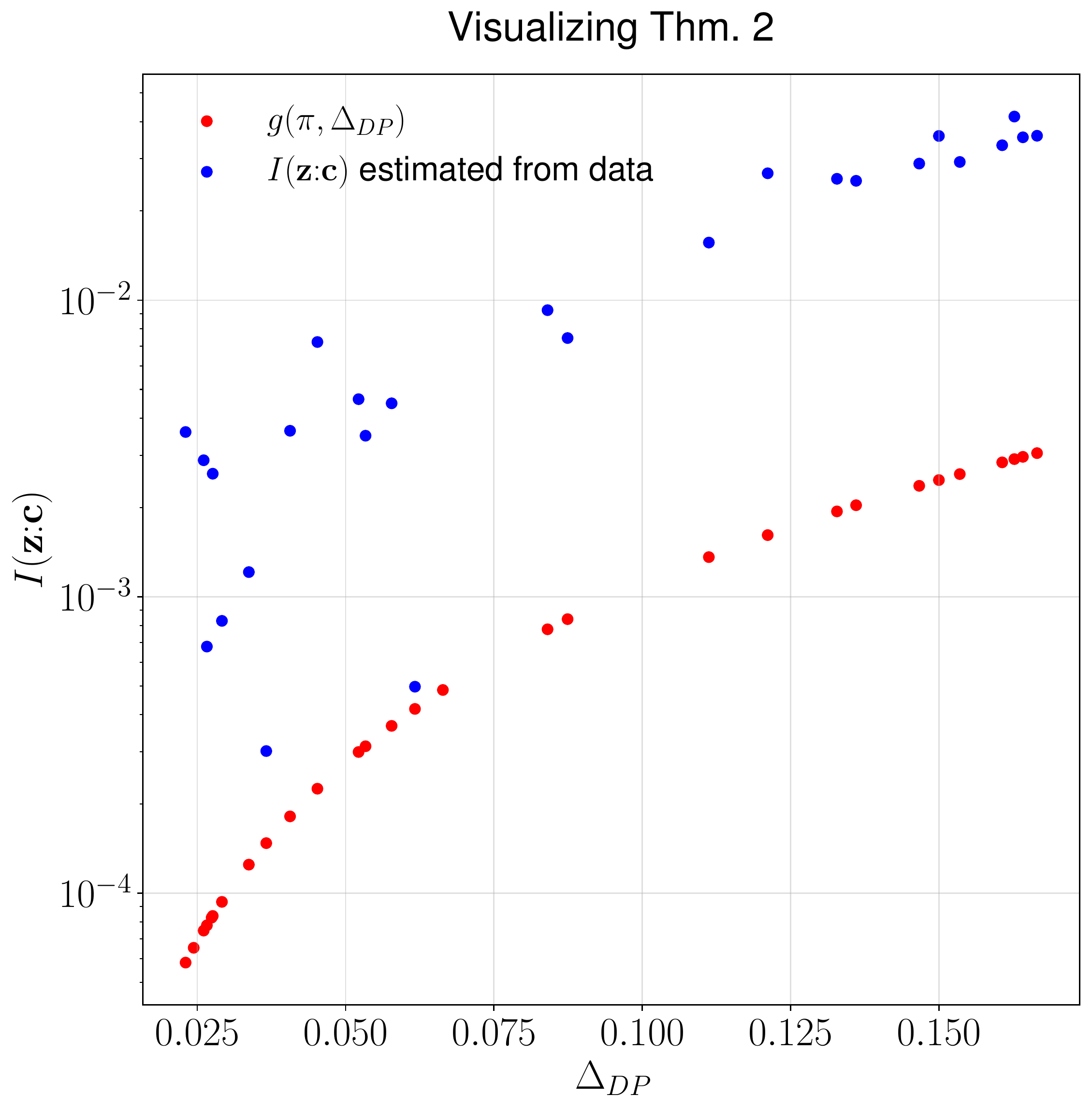}
        \caption{Visualizing Thm.~\ref{thm:parity_mi_relation} for \textit{UCI Adult} dataset. $I(\mbz:\mbc)$ vs. $\DP$ for the representations learned by varying the loss coefficient $\beta$ in the objective of Eq.~\ref{eq:conditional_objective} are shown in blue. Corresponding $g(\pi,\DP)$ is shown in red.}
        \label{fig:mi_parity_bound}
    \end{figure}
    Fig.~\ref{fig:mi_parity_bound} visualizes the bound from Thm.~\ref{thm:parity_mi_relation} for \textit{UCI Adult} dataset. We plot $I(\mbz:\mbc)$ and parity for the representations learned by varying the loss coefficient $\beta$ in the objective of Eq.~\ref{eq:conditional_objective} (shown in blue). This is contrasted with the $I(\mbz:\mbc)$ computed from the Thm.~\ref{thm:parity_mi_relation}, i.e., $g(\pi,\DP)$ (shown in red). We can see that --- a) As predicted by Thm.~\ref{thm:parity_mi_relation}, $I(\mbz:\mbc)$ is always higher than $g(\pi,\DP)$, which is the minimum mutual information predicted due to the observed statistical parity value. b) parity decreases with decreasing $I(\mbz:\mbc)$.

    \subsubsection{When $\mbc$ is multinomial:}
        Letting $\pi_i=P(\mbc=i)$, we can write
        \begin{align*}
            I(\mbz:\mbc)
            &= \sum_{i} \pi_i \KL {p(\mbz\mid\mbc=i)}{p(\mbz)}\\
            &\geq \sum_i \pi_i f\left(V(p(\mbz\mid\mbc=i), p(\mbz))\right)
        \\
            &\geq  f\left(\sum_i \pi_i V(p(\mbz\mid\mbc=i), p(\mbz))\right)
            &\text{(Due to convexity of $f$)}
        \\
            &\geq  f\left(\max_{i,j} \pi_i \|p(\mbz\mid\mbc=i)- p(\mbz)\| + \pi_j \|p(\mbz\mid\mbc=j)- p(\mbz)\|\right)
            &\text{($f$ is strictly increasing)}
        \\
            &\geq  f\left(\max_{i,j} \alpha \|p(\mbz\mid\mbc=i) - p(\mbz\mid\mbc=j)\|\right)
            &\text{(By triangle inequality and increasing nature of $f$)}
        \\
            & =  f\left(\alpha
                \max_{i,j}  \|p(\mbz\mid\mbc=i) - p(\mbz\mid\mbc=j)\|
            \right) \numbereqn \label{eq:multinomial_mi_variational_distance}
        \end{align*}
        where $\alpha=\min_k \pi_k$.

        \noindent We can get a slightly better $\alpha$ by seeing that $\alpha$ can be $\min \{\pi_{i'}, \pi_{j'}\}$, where $i', j' = \argmax_{i,j} \|p(\mbz\mid\mbc=i) - p(\mbz\mid\mbc=j)\|$.

        \noindent Similar to Eq.~\ref{eq:variational_distance_and_parity}, we can show that $
            \|p(\mbz\mid\mbc=i) - p(\mbz\mid\mbc=j)\|
            \geq
            |P(\hat \mby\mid \mbc=i) - P(\hat \mby\mid \mbc=j)|
        $
        and so plugging this in Eq.~\ref{eq:multinomial_mi_variational_distance},
        we get the required result.
        \[
            I(\mbz:\mbc) \geq f\left(\alpha  \DP(\mathcal A, \mbc)\right)
        \]

    \subsection{Proof of Proposition~\ref{prop:conditional_nce}}\label{subsec:cond_nce_proof}
        \begin{proposition} \textbf{(Proposition~\ref{prop:conditional_nce} restated)}
            \label{prop:conditional_nce_restated}
            For any
            $u, v, w\sim p(\mbu, \mbv, \mbw)$,
            $\tilde u \sim p(\mbu\mid\mbw)$,
            and function $f$, we have
            \begin{equation}
                I(\mbu: \mbv\mid\mbw) \geq \mathbb \E_{\mbu,\mbv,\mbw}\log  \frac {e^{f(u,v,w)}} {\frac 1 M \sum_{j=1}^M e^{f(\tilde u_j, v, w)}}
            \end{equation}
            where,
            $u, \tilde u \in \mathcal U$,
            $v \in \mathcal V$,
            $w \in \mathcal W$,
            $f\mathrel{:}\mathcal U \times \mathcal V\times \mathcal W \rightarrow \mathbb R$
            and M is the number of samples from $p(\mbu\mid\mbw)$.
        \end{proposition}

        \begin{proof}
            To prove this, we approximate $p(\mbu\mid\mbv,\mbw)$ with a variational distribution $q$
            \begin{align*}
            q(\mbu\mid\mbv, \mbw) &= \frac {p(\mbu\mid\mbw)e^{g (u,v,w)}}{Z(v,w)}\\
            Z(v,w) &= \int_u du\ {p(\mbu\mid\mbw)e^{f (u,v,w)}} = E_{\mbu\mid \mbw} e^{g(u, v,w)}
            \end{align*}
            then we have
            \begin{align*}
                I(\mbu: \mbv\mid\mbw)
                    & =\E_{\mbu, \mbv,\mbw}\log \cfrac{p(\mbu\mid\mbv,\mbw)}{p(\mbu\mid\mbw)} \\
                    &= \E_{\mbu, \mbv,\mbw}\log \cfrac{q(\mbu\mid\mbv,\mbw)}{p(\mbu\mid\mbw)} + \KL {p(\mbu\mid\mbv,\mbw)}{q(\mbu\mid\mbv,\mbw)}\\
                &\geq  \E_{\mbu, \mbv,\mbw} g(u,v,w) - \E_{\mbv,\mbw} \log Z(v,w)\\
                &\geq  \E_{\mbu, \mbv,\mbw} g(u,v,w) - \E_{\mbv,\mbw} \log \E_{\mbu\mid \mbw} e^{g(u,v,w)}\\
                &\geq  \E_{\mbu, \mbv,\mbw} g(u,v,w) - \E_{\mbv,\mbw} \E_{\mbu\mid \mbw} e^{g(u, v,w)}+1
            \end{align*}
            Last inequality is due to $\log x \leq x-1$.
            Similar to~\citet{poole2019variational}, we assume that we have $M-1$ extra samples from $p(\mbu|\mbw)$, and let $g(u,v,w) = g(u_{1:M},v,w) = f(u_1, v, w) - \log {\frac 1 M \sum_{i=1}^M e^{f(u_i, v,w)}}$, where for notational ease $u_1=u$ and $u_{2:M}$ are the additional samples.
            \noindent With these choices, we have:
            \begin{align*}
                E_{\mbv,\mbw, \mbu\mid \mbw} e^{g(u, v,w)} &= E_{\mbv,\mbw, \mbu\mid \mbw} \frac {e^{f(u, v,w)}}{\frac 1 M \sum_{i=1}^M e^{f(u_i, v,w)}}\\
                &=E_{\mbv,\mbw,\mbu\mid \mbw} \frac {\frac 1 M \sum_{j=1}^M e^{f(u_i, v,w)}}{\frac 1 M \sum_{i=1}^M e^{f(u_i, v,w)}} = 1
                &\text{(replacing with M sample mean estimate)}
            \end{align*}
            and the result follows.
        \end{proof}

\section{On Maximizing $I(\mby:\mbz)$ vs $I(\mby:\mbz|\mbc)$}\label{sec:synergy}
As discussed in Sec.~\ref{subsec:interference} and demonstrated experimentally in Fig.~\ref{fig:cond_vs_non_cond}, we find that maximizing $I(\mby:\mbz|\mbc)$ is a better objective than maximizing $I(\mby:\mbz)$. However, it may perform poorly in specific scenarios.

One such extreme case is when $\mbx$ and $\mbc$ together are sufficient to predict $\mby$, but neither can predict $\mby$ by itself. Mathematically, that means $\mbx \perp \mby$, $\mbc \perp \mby$, and $H(\mby|\mbx, \mbc)=0$. When $\mbx,\mbc,\mby$ are 1-dimensional binary random variables, these conditions are met by the XOR function, i.e., $y=x\oplus c$. In this case, if $z$ is a function of just $x$, the optimal solution to both objectives~\ref{eq:objective} and~\ref{eq:conditional_objective} will be $z=x$. As a result, the representations $z$ are not useful to predict $y$. However, suppose z is a function of both $x$ and $c$. In that case,  objective~\ref{eq:objective} may produce representations predictive of $c$, depending on the multiplier $\beta$. However, for objective~\ref{eq:conditional_objective}, the optimal representation will still be $z=x$, regardless of the coefficient $\beta$, and hence it may fail to learn any useful representations at all. However, in this scenario, one may also question the need to use a fair representation learning framework as $\mbx$ is already independent of $\mbc$. A perfect classifier, without any fairness constraints, will already produce outcomes with 0 statistical parity. Our choice of using $I(\mby:\mbz | \mbc)$ is justified by the empirical performance shown in Sec.~\ref{sec:experiments}.

\section{Training and Evaluation Details}\label{sec:training_details}

\subsection[]{Loss function and FCRL Implementation}\label{subsec:loss_implementation}
Using Eq.~\ref{eq:mi_diff}, we write objective~\ref{eq:conditional_objective} as:
\begin{align*}
    I(\mby:\mbz|\mbc)- \beta I(\mbz:\mbc) &  =I(\mby:\mbz|\mbc)- \beta\left(I(\mbz:\mbx)-I(\mbx:\mbz|\mbc)\right)\\
    &= I(\mby:\mbz|\mbc) + \beta I(\mbx:\mbz|\mbc) - \beta I(\mbz:\mbx)\\
\end{align*}
We lower bound $I(\mby: \mbz|\mbc)$ and $I(\mbx:\mbz|\mbc)$ using bounds from Eq.~\ref{eq:lower_bound_conditional_classifier} \&~\ref{eq:contrastive_estimation}, respectively. $I(\mbz:\mbx)$ is upper bounded using Eq.~\ref{eq:rate}. Introducing parameters $\phi$ and $\psi$, we have:
\begin{align}
    & I(\mby:\mbz|\mbc)- \beta I(\mbz:\mbc) \nonumber \\
    &\geq H(\mby)+\E_{\mby,\mbz, \mbc} r(\mby|\mbz, \mbc;\psi)
    + \beta \E_{\mbz,\mbx,\mbc} \left[f(z,x,c)-\log \cfrac 1 M \sum_{j=1}^M e^{f(\tilde z_j, x,c)}\right]
    -\beta \E_\mbx \KL{q(\mbz|\mbx;\phi)}{p(\mbz)}
\end{align}
$H(\mby)$ is a constant.  As $\mby$ is a binary variable, $\E_{\mby,\mbz, \mbc} r(\mby|\mbz, \mbc;\psi)$ is implemented as the binary cross-entropy loss. Similar to \citet{cpc}, we use a bilinear function to parametrize $f$  and
$$\exp\{{f(z, x, c)}\} = \texttt{soft-plus}\left[(\mbW_z z)^T\mbW_c^T e(x;\theta')\right]$$
where $\mbW_z, \mbW_c, \theta'$ are learnable parameters.
We let $q(\mbz|\mbx;\phi)$ be a diagonal gaussian distribution whose mean and variance are computed via a neural network  parametrized by  $\phi$, and $p(z)\sim  \normal(\mathbf{0},\mathbf I) $. KL divergence term in the last expression can be thus computed in the closed form.
\begin{equation}
    \E_\mbx \KL{q(\mbz|\mbx;\phi)}{p(\mbz)} =
    \E_\mbx \sum_{k=1}^d\cfrac 1 2  \left(-2\log \sigma(x)_k + {(\sigma(x)_k+\mu(x)_k)^2}- 1 \right)
\end{equation}
where $z$ is d-dimensional and $\mu(x)_k, \sigma(x)_k$ denotes the mean and variance of the $k^{th}$ component. Next, we discuss the  implementation of this objective for a batch.

\begin{center}
\begin{minipage}{0.95\textwidth}
\newcommand\mycommfont[1]{\footnotesize\textcolor{blue}{#1}}
\SetCommentSty{mycommfont}
\begin{algorithm}[H]
    \SetAlgoLined
    \SetNoFillComment
    \LinesNotNumbered
    \DontPrintSemicolon
    \KwData{Batch data $(X, Y, C)$}
    \KwResult{loss}

    \;
    \tcc{encode input to get distribution parameters, $q(\mbz|\mbx;\phi)$ and sample}
    $\mu(X), \sigma(X) = NN(X;\phi)$\;
    $\epsilon$ = \texttt{numpy.random.randn}(B, d)\tcp*{$\epsilon\sim \mathcal N (0, \mathbf I)$}
    $Z = \mu(X) + \epsilon*\sigma(X)$\;

    \;
    \tcc{label loss}
    $\hat Y = NN(Z,C;\psi)$\;
    label-loss = cross-entropy($\hat Y, Y$)\;

    \;
    \tcc{$I(\mbz:\mbx)$ term}

    rate = $\cfrac 1 {2B} \sum_{i=1}^B \left\{ \sum_{k=1}^d -2\log \sigma(X_i)_k + (\sigma(X_i)_k+\mu(X_i)_k)^2-1\right\}$\;
    \;

    \tcc{contrastive loss computation}
    \tcp{transformations}
    $Z' = NN(X;\theta')$\;
    $Z'' =\mbW_z Z$\;
    \tcp{iterate over each sample and compute contrastive loss}
    \For { i in $\{0\ldots B-1\}$}{
        $\tilde Z'' = \{Z''_j \mathbin{:} c_j = c_i\}$ \tcp*{$\tilde Z''$ is matrix of all $Z''_j$ such that $c_i=c_j$}
        $M = |\{Z''_j \mathbin{:} c_j = c_i\}|$\;
        $c=c_i$\;
        contrastive-loss += $ \cfrac 1 B \left(\log (\texttt{soft-plus}(Z^{''T}_i \mbW_{c}^T Z'_i)) - \log \left\{\cfrac 1 M \sum_j \texttt{soft-plus}(\tilde Z^{''T}_j\mbW_{c}^T Z'_i)\right\}\right)$\;
    }
    loss = $\text{label-loss} + \beta\times \text{contrastive-loss} -\beta \times \text{rate}$

    \caption{Computation steps for objective~\ref{eq:conditional_objective}.
    \protect \\\hspace{\textwidth}B is the batch size. $NN(\cdot;\theta)$ denotes some neural network parametrized by $\theta$. $X_i$ denotes $i^{th}$ sample from the batch $X$. The implementation is available at \codeurl.
    }
    \end{algorithm}
\end{minipage}
\end{center}

\subsection[]{Architecture Details}
\newcommand{\relu}{ReLU }
Table~\ref{tab:components} provides details about the architecture components used in each method.
Table~\ref{tab:arch_hidden_layer_size} provides details about each component's hidden layer dimension for the \textit{UCI adult} and \textit{Heritage Health} dataset. We set $d=8$ as the representation size for each method and use one hidden layer neural network with \relu non-linearity for all the components. We train all the models for 200 epochs with Adam optimizer and learning rate $10^{-3}$ except Adversarial Forgetting, which was trained with $10^{-4}$ learning rate as prescribed in~\citet{jaiswal2020invariant}. We implement FCRL, CVIB, Adversarial Forgetting. and MaxEnt-ARL in PyTorch~\cite{pytorch}. We used official implementations of MIFR\footnote{\url{https://github.com/ermongroup/lag-fairness}} and LAFTR\footnote{\url{https://github.com/VectorInstitute/laftr}}. CVIB, MIFR, and our approach use stochastic representations, which are  distributed as gaussian with mean and variance generated by the encoder. Other methods use deterministic representations. As the inputs are between $0$ and $1$, we use  cross-entropy loss as the reconstruction loss wherever a decoder is used.

\begin{table}
    \centering
    \begin{tabular}{l c c c c c c}
        \toprule
        Method              & Encoder           & Decoder           & Discriminator     & Predictor         & Mask-Encoder      & $e$\\
        \cmidrule(l){1-1}   \cmidrule(lr){2-2}  \cmidrule(lr){3-3}  \cmidrule(lr){4-4}  \cmidrule(lr){5-5}  \cmidrule(lr){6-6}  \cmidrule(lr){7-7}
        FCRL (Ours)         &
                                \cmark          & \xmark            & \xmark            & \cmark            & \xmark             & \cmark\\
        CVIB~\cite{moyer2018invariant} &
                                \cmark          & \cmark            & \xmark            & \cmark            & \xmark             & \xmark\\
        MIFR~\cite{song2019learning} &
                                \cmark          & \cmark            & \cmark            & \xmark            & \xmark             & \xmark\\
        LAFTR~\cite{madras2018learning} &
                                \cmark          & \xmark            & \cmark            & \cmark            & \xmark             & \xmark\\
        Adversarial Forgetting~\cite{jaiswal2020invariant} &
                                \cmark          & \cmark            & \cmark            & \cmark            & \cmark             & \xmark\\
        MaxEnt-ARL~\cite{roy2019mitigating} &
                                \cmark          & \xmark            & \cmark            & \cmark            & \xmark             & \xmark\\
        \bottomrule
    \end{tabular}
    \caption{Architecture components in different methods. FCRL, CVIB, and MIFR use stochastic embeddings that are gaussian distributed whose mean and variance are generated by an encoder. Adversarial Forgetting uses a separate encoder for the drop-out mask, called as mask-encoder in this table. Our method uses a neural network, $e$ to learn the contrastive function for mutual information estimation (Eq.~\ref{eq:contrastive_estimation}). }\label{tab:components}
\end{table}
\begin{table}
    \centering
    \begin{tabular}{l c c}
        \toprule
        Component         & UCI Adult         & Heritage Health \\
        \cmidrule(l){1-1} \cmidrule(lr){2-2}  \cmidrule(lr){3-3}
        Encoder           & 50                & 100 \\
        Decoder           & 50                & 100 \\
        Predictor         & 50                & 50  \\
        Discriminator     & 50                & 50  \\
        Mask-Encoder      & 50                & 100 \\
        $e$               & 50                & 100 \\
        \bottomrule
    \end{tabular}
    \caption{Hidden layer dimensions of different components for each dataset. We use the same architecture for components in each method. We use one hidden layer neural network and \relu non-linearity in all cases.}\label{tab:arch_hidden_layer_size}
\end{table}

\subsection{Hyperparameter Grids to Generate Parity-Accuracy Curves}
\label{subsec:hparam_grid}
Every method has different number of parameters to vary and they may require different ranges. Therefore, we have a different number of models or points on the parity-accuracy curve for each method. For our results, we use the following settings.

    \paragraph{FCRL (Ours)}
    During the initial experiments, we varied $\lambda$ in $\{0.2, 0.5, 1.0, 2, 5\}$  and $\beta$ is varied from 0.01 to 0.1 in incremental steps of 0.01, and 0.1 to 1.0 in incremental steps of 0.1.
    For final results, we fix $\lambda=2$ and $\beta$ is varied from 0.005 to 0.05 in incremental steps of 0.005 and 0.05 to 1.0 in incremental steps of 0.05.
    For Fig.~\ref{fig:cond_vs_non_cond} and~\ref{fig:contrastive_vs_reconstruction}, we used $\lambda=1$ and $\beta$ same as initial experiments.

    \paragraph{Adversarial Forgetting}
    \citet{jaiswal2020invariant} do not provide the range of loss coefficients. Therefore, we searched over a reasonable range.
    Adversarial Forgetting has three loss coefficients $\rho, \delta$, and $\lambda$. $\delta$ which is the coefficient of discriminator is chosen from the set $\{0.001, 0.005, 0.01, 0.05, 0.1, 0.5, 1.0, 5.0, 10.0, 50.0\}$. We varied $\rho$ and $\lambda$, coefficient of reconstruction, and mask regularization in $\{0.001, 0.01, 0.1\}$. All the combinations of $\rho, \delta$, and $\lambda$ were tried leading to 90 different models. We update discriminator 10 times for every update of all the other modules as mentioned by the authors.

    \paragraph{MaxEnt-ARL}
    This method has only one coefficient $\alpha$, which is varied within $\{0.1, 0.2, 0.5, 1, 2, 5, 10, 20, 50, 100\}$. We update discriminator only once for every update of other modules.

    \paragraph{LAFTR}
    This method has a single coefficient $\lambda$, varied within $\{0.1, 0.2, 0.3, 0.5, 0.7, 1.0, 1.5, 2.0, 3.0, 4.0\}$ as prescribed by the authors.

    \paragraph{CVIB}
    \citet{moyer2018invariant} designed their approach keeping in mind invariance and  did not prescribe a suitable range for the coefficients. Their approach has two coefficients $\lambda$ -- the coefficient of $I(\mbz:\mbc)$, and $\beta$ -- the coefficient of information bottleneck $I(\mbz:\mbx)$. We choose $\beta$ from $\{0.001, 0.01, 0.1\}$ and $\lambda$ is varied from $0.01$ to $0.1$ in steps of 0.01 and $0.1$ to $1.0$ in steps of $0.1$. We observed that the representations learned with lower values of $\beta$ do not achieve lower \DP.

    \paragraph{MIFR}
    For MIFR, we vary loss coefficients $e_1$ and $e_2$ in the range prescribed in the original work, i.e., $e_1$ is varied from $\{0.0, 0.2, 0.1, 1.0, 2.0, 5.0\}$ and $e_2$ is varied from $\{0.1, 0.2, 1.0, 2.0, 5.0\}$.

\subsection[]{Evaluation Details}
    Following steps are involved in the evaluation of any method:
    \begin{itemize}
        \item \textbf{Learn Representations / Encoder:} We train each method on the train set to learn the encoder. For both the datasets, we train separate models with different loss coefficients  described in Sec.~\ref{subsec:hparam_grid}.
        \item \textbf{Generate Representations:} After learning the encoder, we generate representations for each $x$ in the test and train set.
        \item \textbf{Evaluation via Downstream Classification Task:}
        We simulate the scenario of downstream classification task to evaluate representations
        which involves the following steps. Train, test, validation set below refer to the representations of the train, test and validation set generated by the encoder. 
        \begin{itemize}
            \item \textit{Preprocessing:} In the preprocessing step, we scale the test and train set using standard scaling. For scaling, the statistics are computed from the train set only. We leverage scikit-learn's preprocessing module for this. Any other preprocessing is valid too.
            \item \textit{Fit a classifier on the train set:} We fit a classifier to the train set. Some classifiers (for instance, MLP) may use early stopping, and therefore, need a validation set. In such cases, the validation set is derived from the train set  by randomly splitting the train set, and keeping 20\% of the train set for validation.
            \item \textit{Evaluation:}  We use the test set to evaluate the \DP\ and accuracy of representations. The test set is used only to evaluate the representations.
        \end{itemize}
    \end{itemize}

    \noindent
    \textbf{Classifier Settings:}
    We use classifiers implemented in scikit-learn~\cite{scikit-learn} to classify representations. We keep the default settings for Logistic regression and Random Forest classifier. For 1- \& 2-hidden-layer MLP, we use 50-dimensional hidden layer with \relu non-linearity and set max epochs to 1000, and patience for early stopping is set to 10 epochs. To report results with SVM, we use an ensemble of 10 SVC classifiers, each trained with different samples. We do not fix seeds in any of our experiments.
\section{Parity vs.\ Accuracy With Different Classification Algorithms}\label{sec:other_eval}
\begin{figure}[]
    \centering
    \begin{subfigure}[]{0.47\textwidth}
        \includegraphics[width=\textwidth]{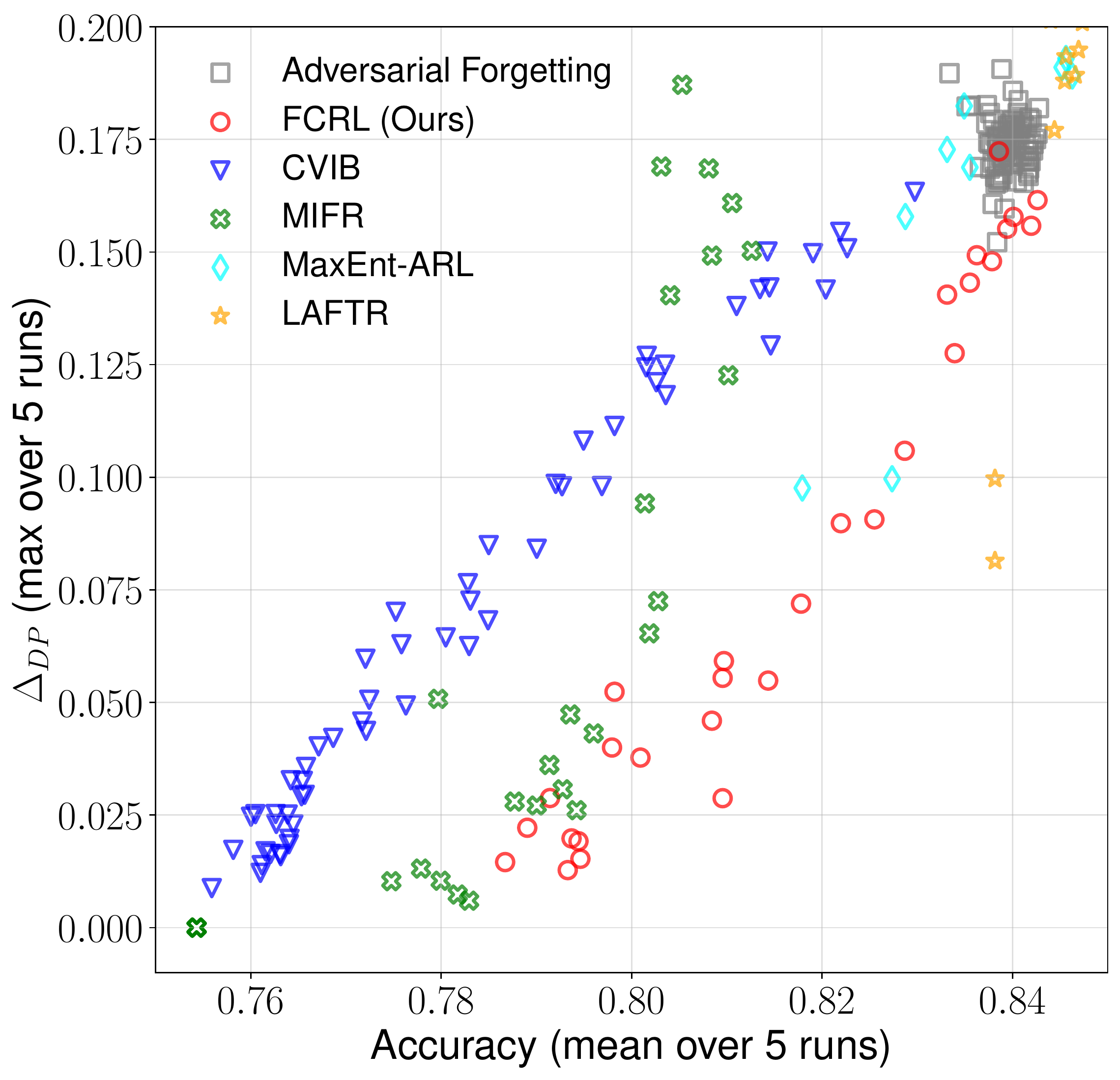}
        \caption{Logistic Regression}
    \end{subfigure}
    \quad
    \begin{subfigure}[]{0.47\textwidth}
        \includegraphics[width=\textwidth]{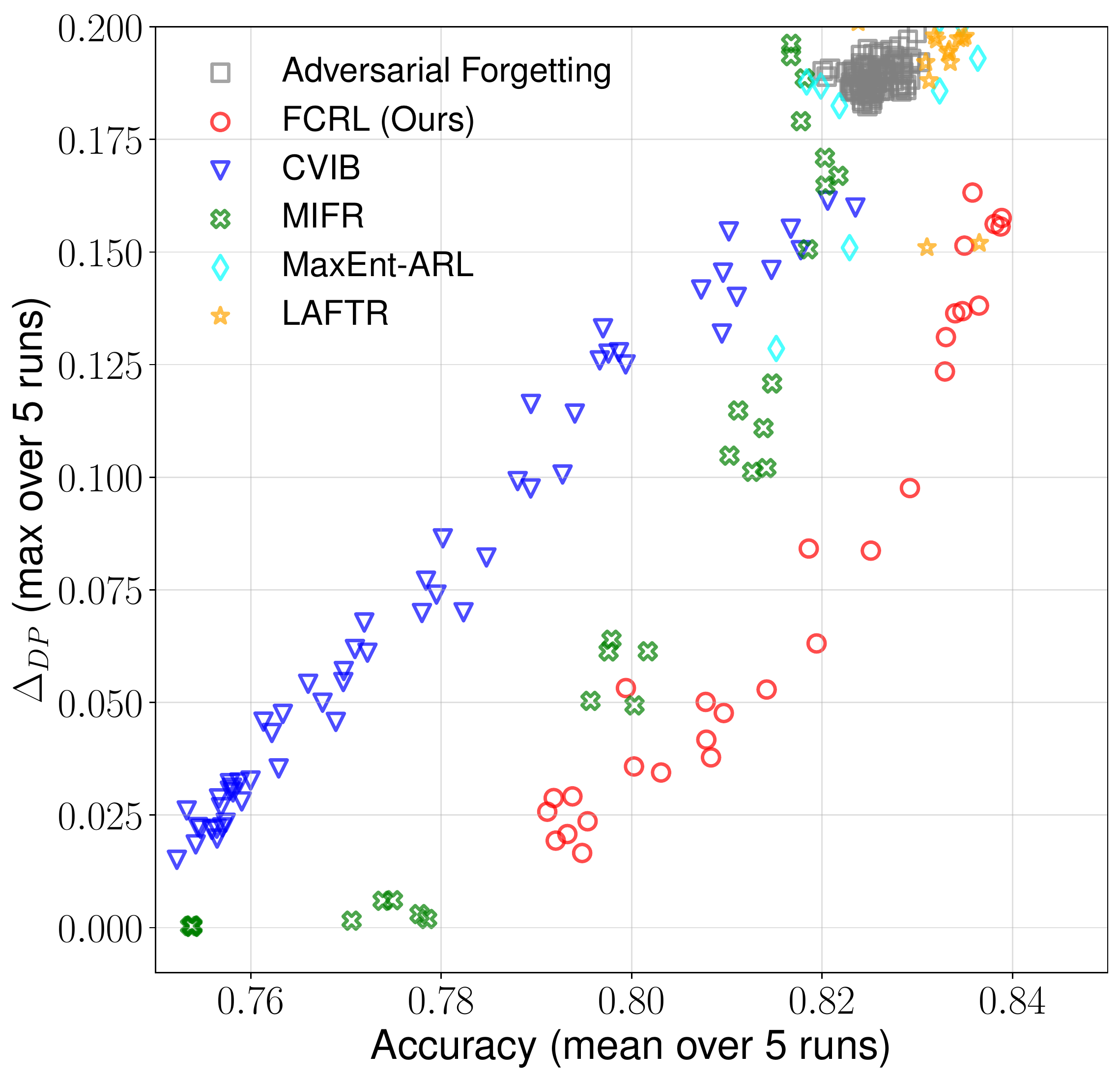}
        \caption{Random Forest}
    \end{subfigure}

    \begin{subfigure}[]{0.47\textwidth}
        \includegraphics[width=\textwidth]{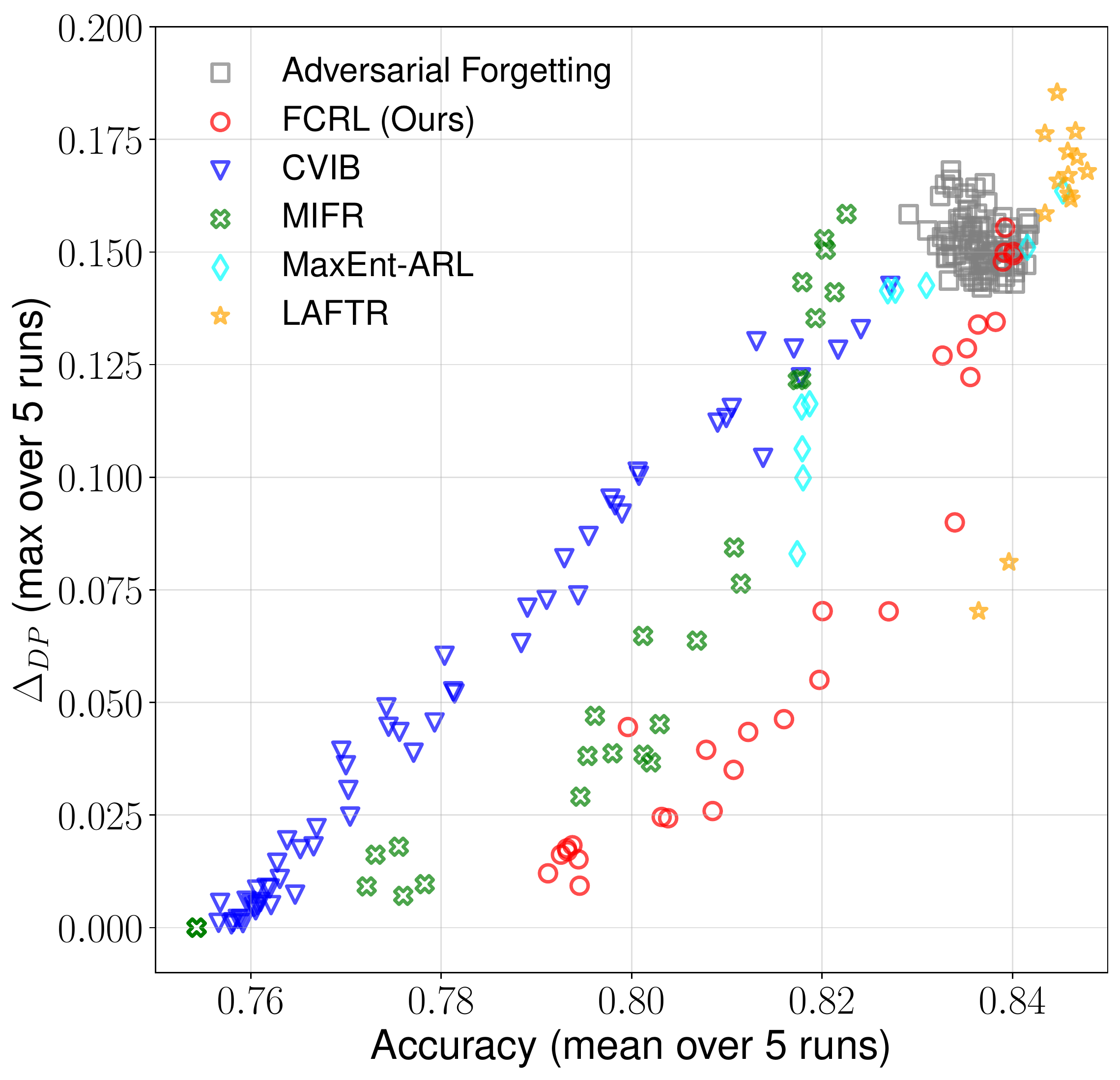}
        \caption{SVM}
    \end{subfigure}
    \quad
    \begin{subfigure}[]{0.47\textwidth}
        \includegraphics[width=\textwidth]{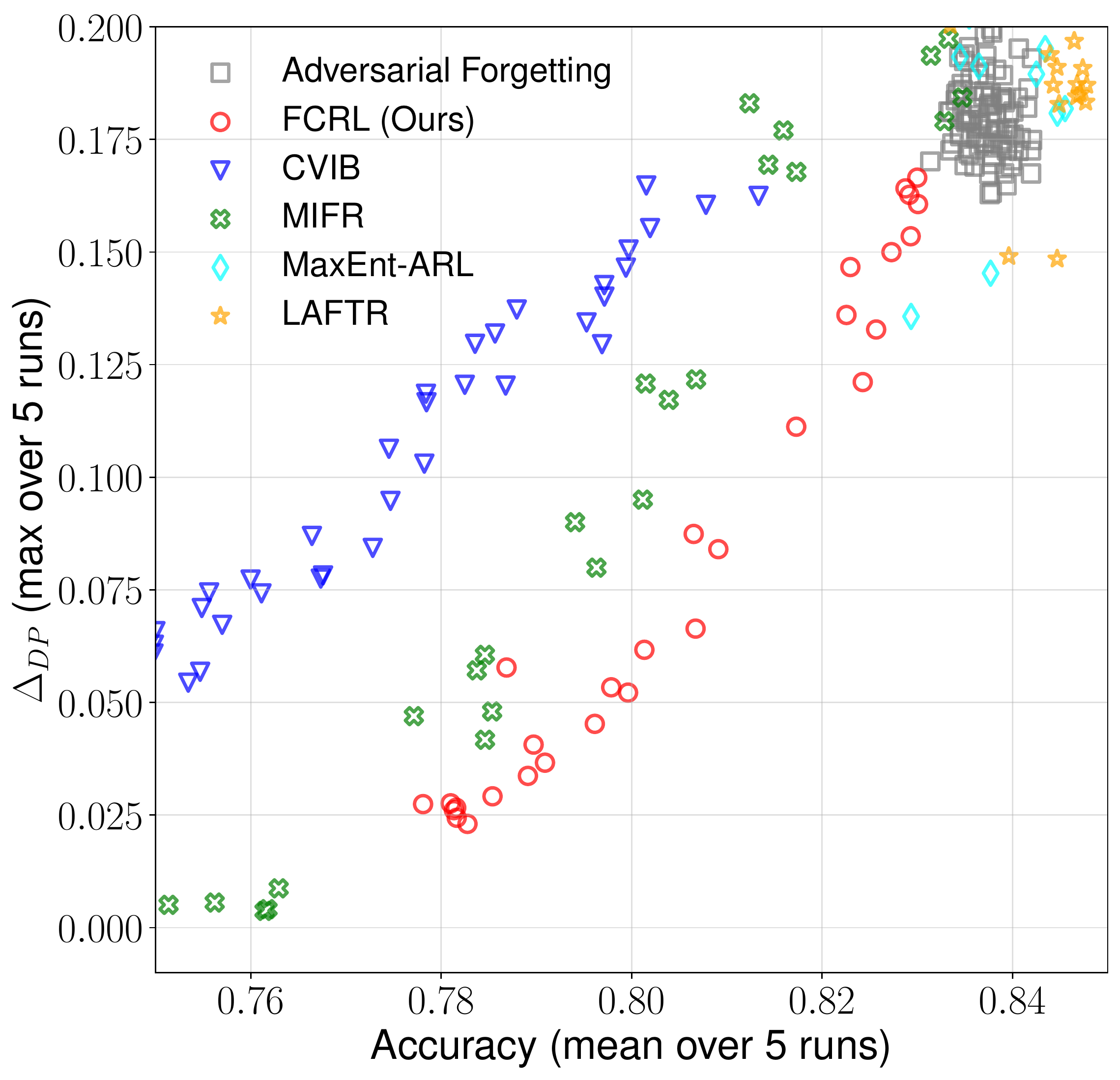}
        \caption{2 Layer MLP}
    \end{subfigure}
    \caption{Parity vs. Accuracy with different classification algorithms for \textit{UCI Adult} dataset.}\label{fig:parity-accuracy-adult}
\end{figure}

\begin{figure}[]
    \centering
    \begin{subfigure}[]{0.47\textwidth}
        \includegraphics[width=\textwidth]{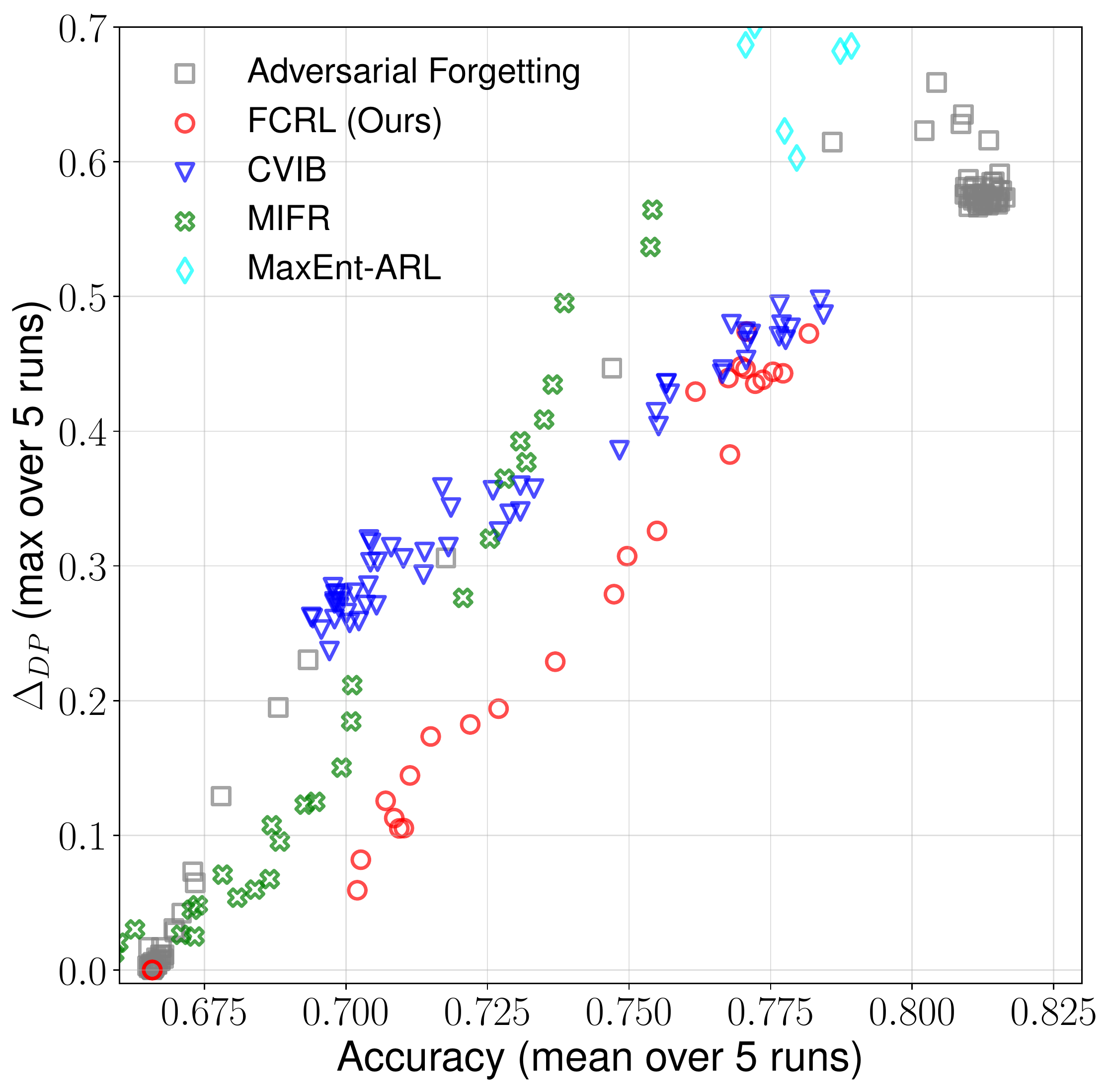}
        \caption{Logistic Regression}
    \end{subfigure}
    \quad
    \begin{subfigure}[]{0.47\textwidth}
        \includegraphics[width=\textwidth]{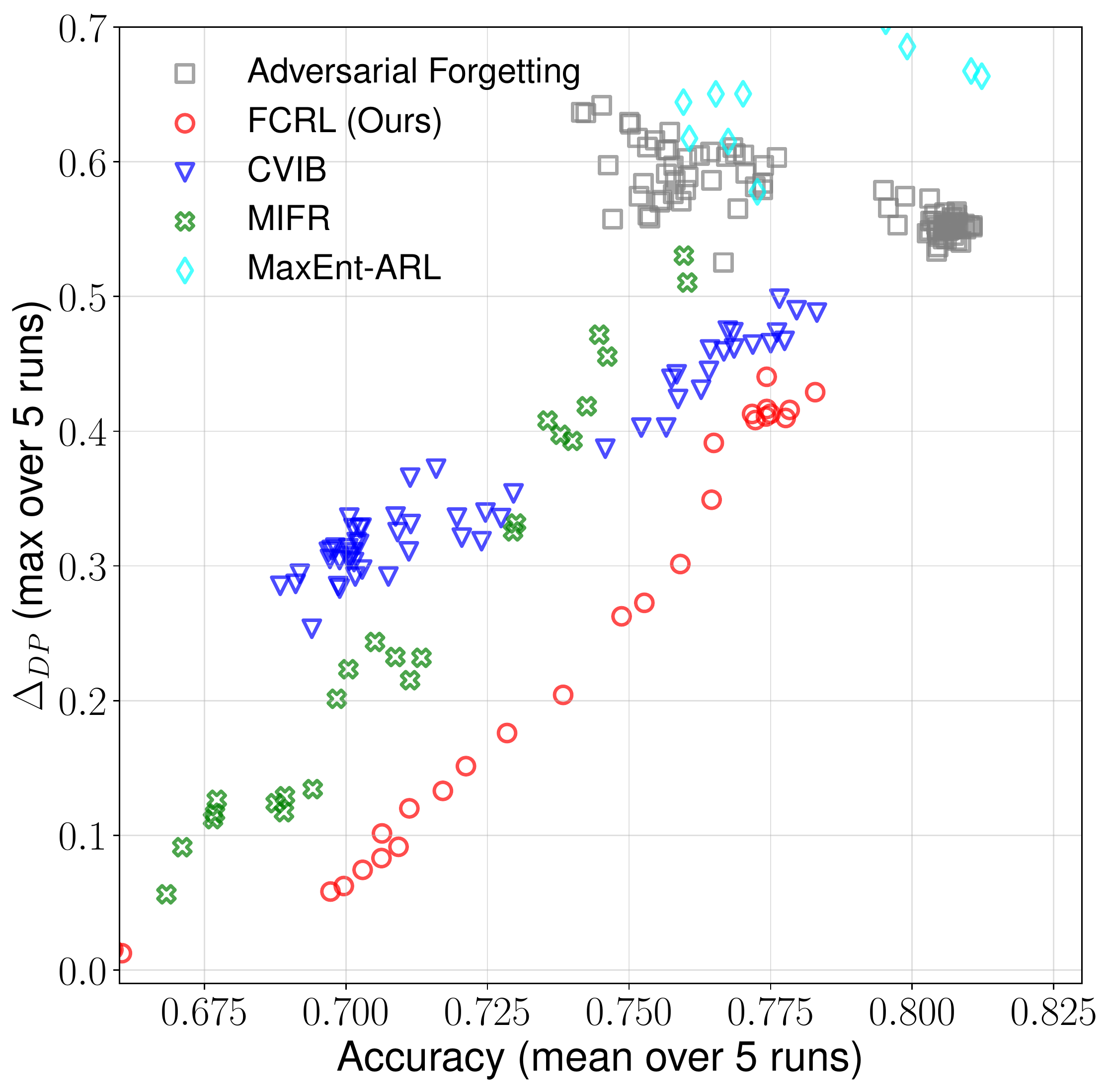}
        \caption{Random Forest}
    \end{subfigure}

    \begin{subfigure}[]{0.47\textwidth}
        \includegraphics[width=\textwidth]{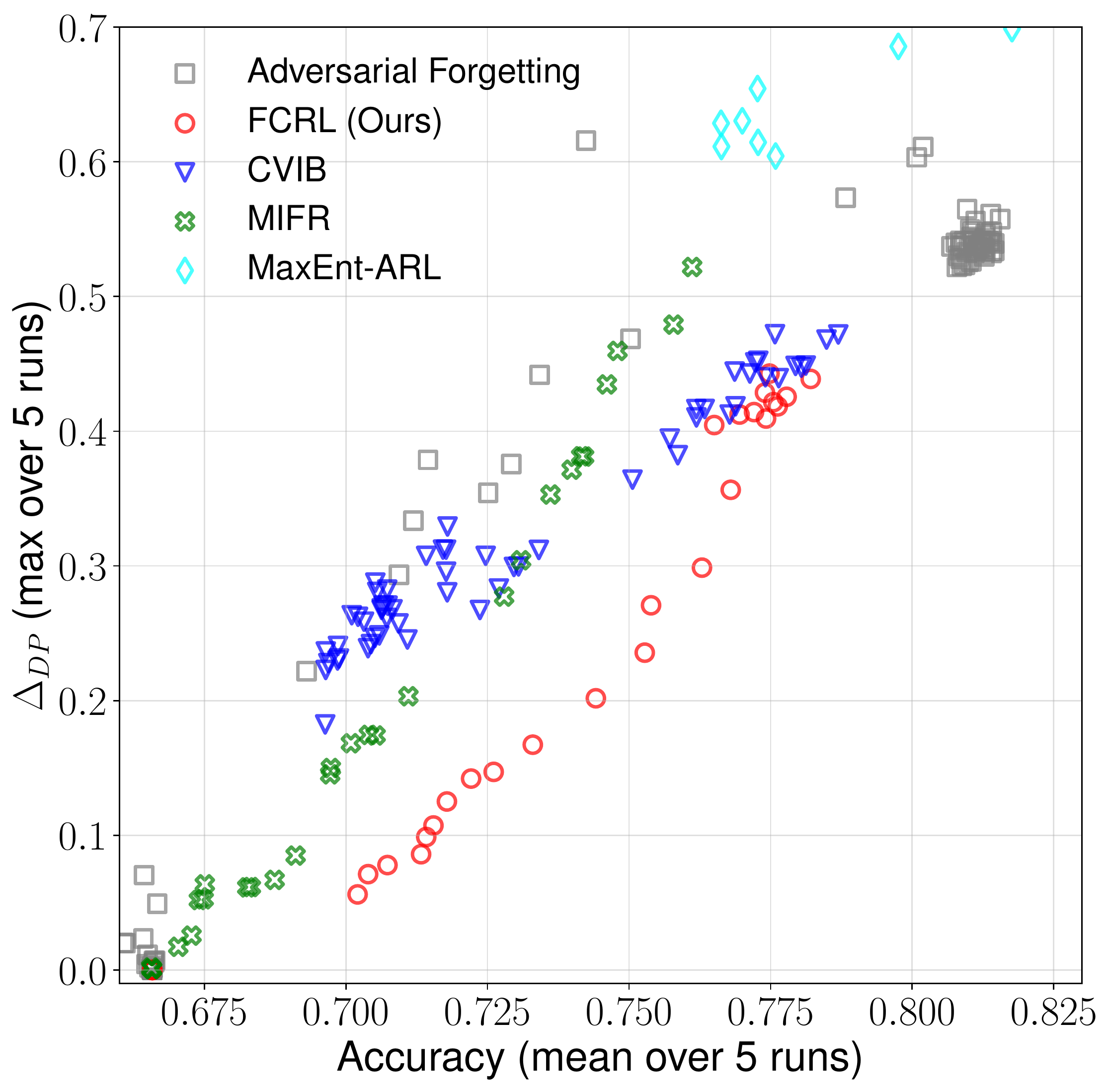}
        \caption{SVM}
    \end{subfigure}
    \quad
    \begin{subfigure}[]{0.47\textwidth}
        \includegraphics[width=\textwidth]{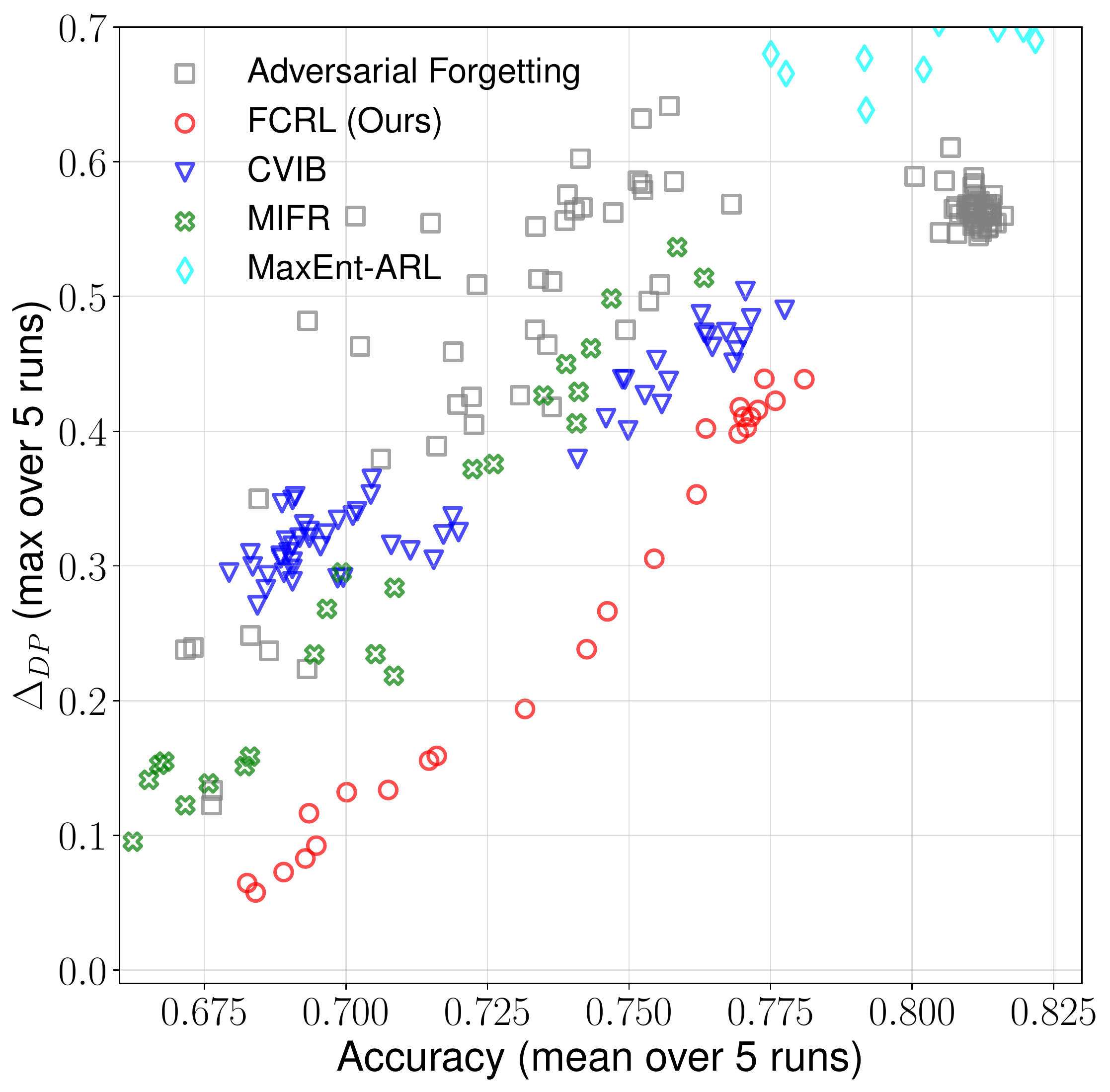}
        \caption{2 Layer MLP}
    \end{subfigure}
    \caption{Parity vs. Accuracy with different classification algorithms for \textit{Heritage Health} dataset.}\label{fig:parity-accuracy-health}
\end{figure}

The area over curve for all methods is shown in Table~\ref{tab:AOPAC_better}.

\noindent
\textbf{UCI Adult:}
Fig.~\ref{fig:parity-accuracy-adult} shows the parity-accuracy curve with different classification algorithms. As we can see, adversarial forgetting and MaxEnt-ARL do not achieve lower \DP\ with any classification algorithm. A couple of models trained with LAFTR  achieved good results (lower \DP) when logistic-regression or SVM is used. However, more complex non-linear models like MLP or random forests make unfair predictions with these representations. MIFR is a method that makes use of adversarial learning along with a loose upper bound on $I(\mbz:\mbc)$. The loose upper bound may not be very helpful in enforcing fairness constraints. We can see from the results of 2-layer MLP that some of the representations that had lower \DP\ with other classification methods  get higher  \DP.
Our method works by minimizing an upper bound of $I(\mbz:\mbc)$ and is not very sensitive to the downstream classifier.

\noindent
\textbf{Heritage Health:}
Fig.~\ref{fig:parity-accuracy-health} shows the parity-accuracy curve with different classification algorithms. MaxEnt-ARL could not achieve lower \DP, whereas we see that adversarial forgetting could achieve lower \DP\ with simple classifiers like logistic regression or SVM. With MLP too,  it is able to achieve lower \DP\, albeit with not very good accuracy. However, results with random forest do not achieve low parity.
Other methods are consistent and do not depend on the complexity of downstream classifiers.

\section{Hidden Information in  Adversarial Representation Learning}
\label{sec:hidden_bias}

\begin{figure}[h!]
    \centering
    \begin{subfigure}[]{0.47\textwidth}
        \includegraphics[width=\textwidth]{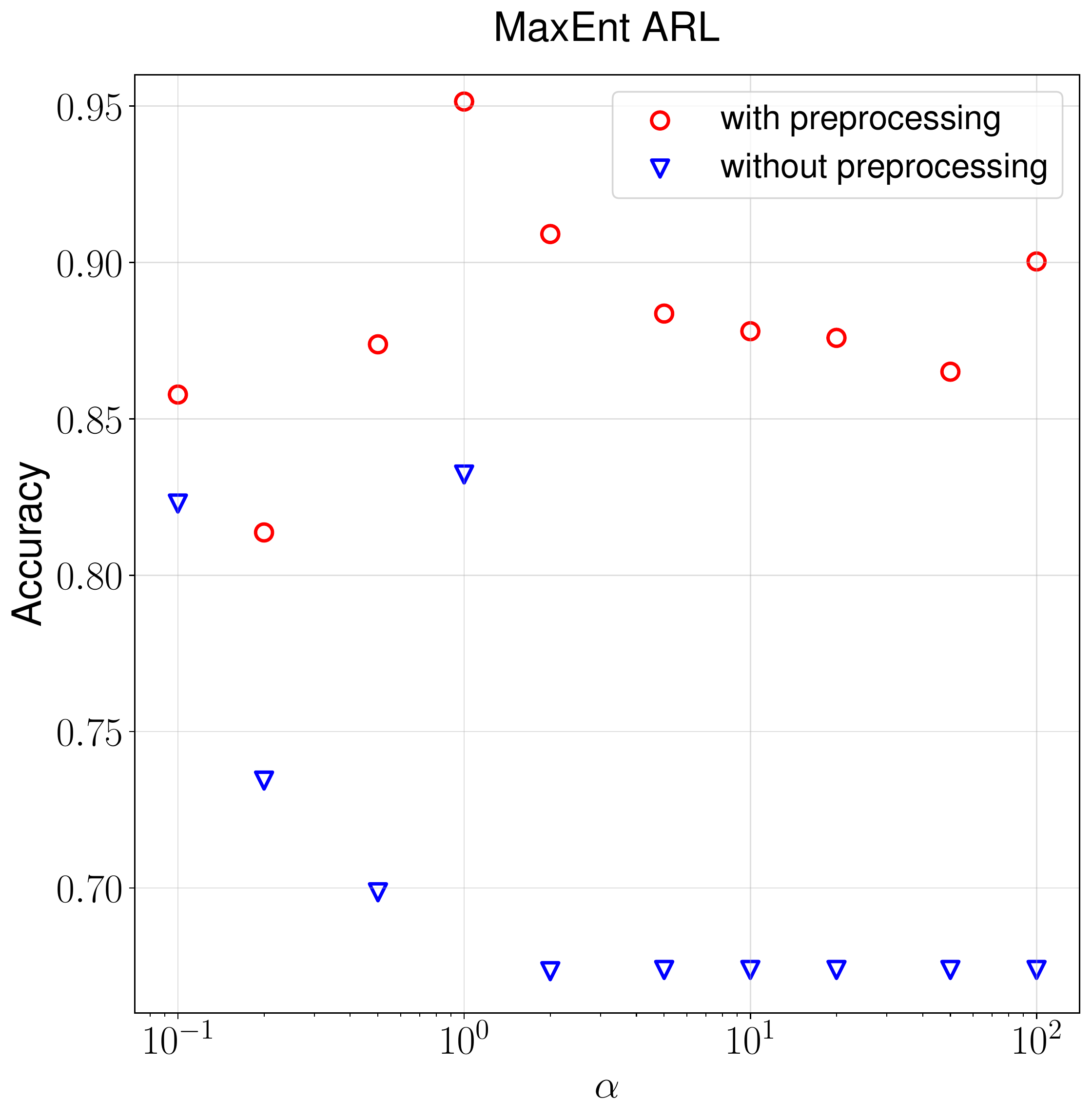}
        \caption{Accuracy of predicting $c$ from $z$ with and without pre-processing as a function of loss paramter $\alpha$. Adversary architecture didnot have batch-norm.}\label{fig:invariance_maxent}
    \end{subfigure}
    \quad
    \begin{subfigure}[]{0.47\textwidth}
        \includegraphics[width=\textwidth]{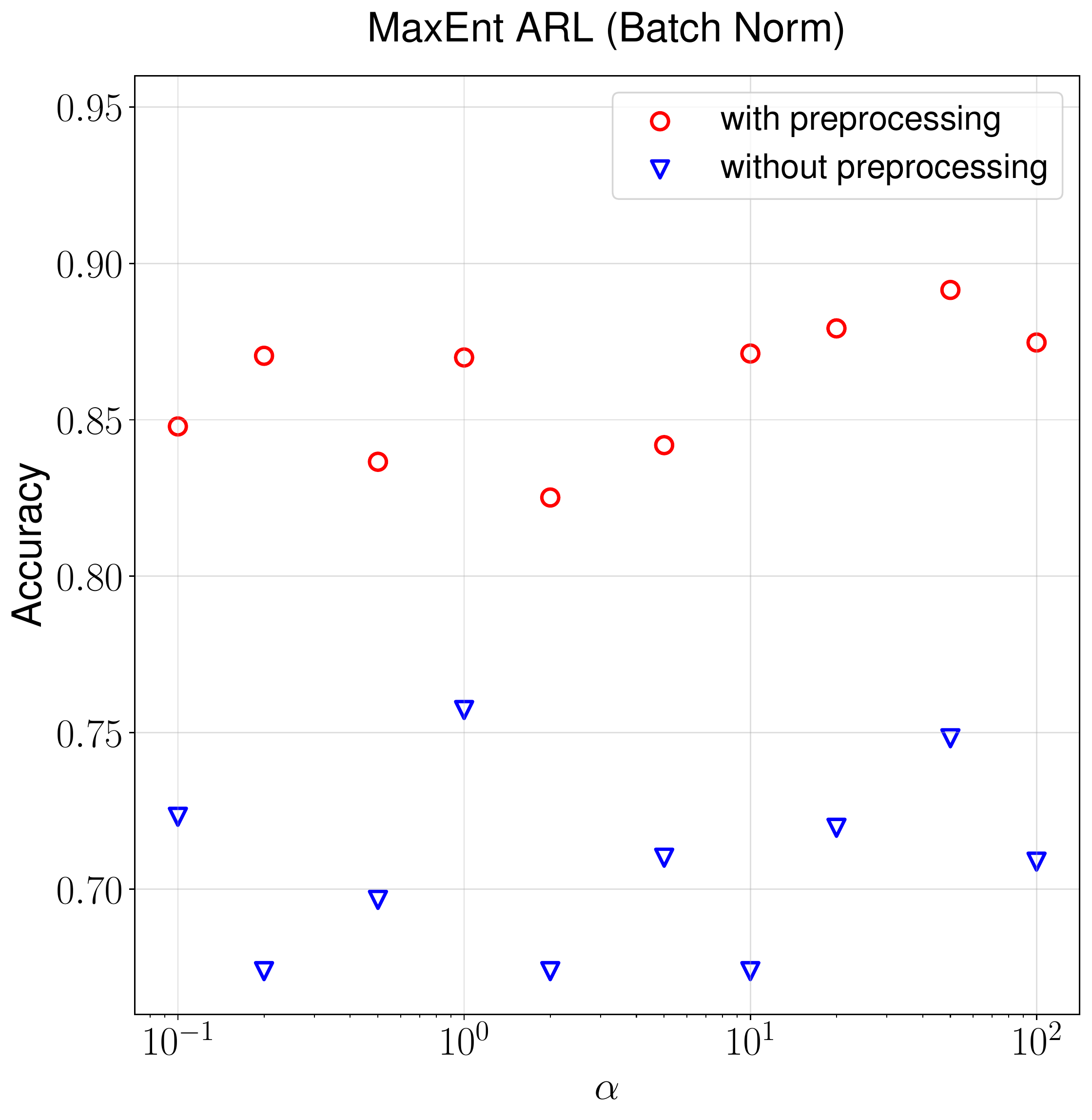}
        \caption{Accuracy of predicting $c$ from $z$ with and without pre-processing as a function of loss paramter $\alpha$. Models were trained with batch-norm in the adversary,}\label{fig:invariance_maxent_bn}
    \end{subfigure}

    \begin{subfigure}[]{0.47\textwidth}
        \includegraphics[width=\textwidth]{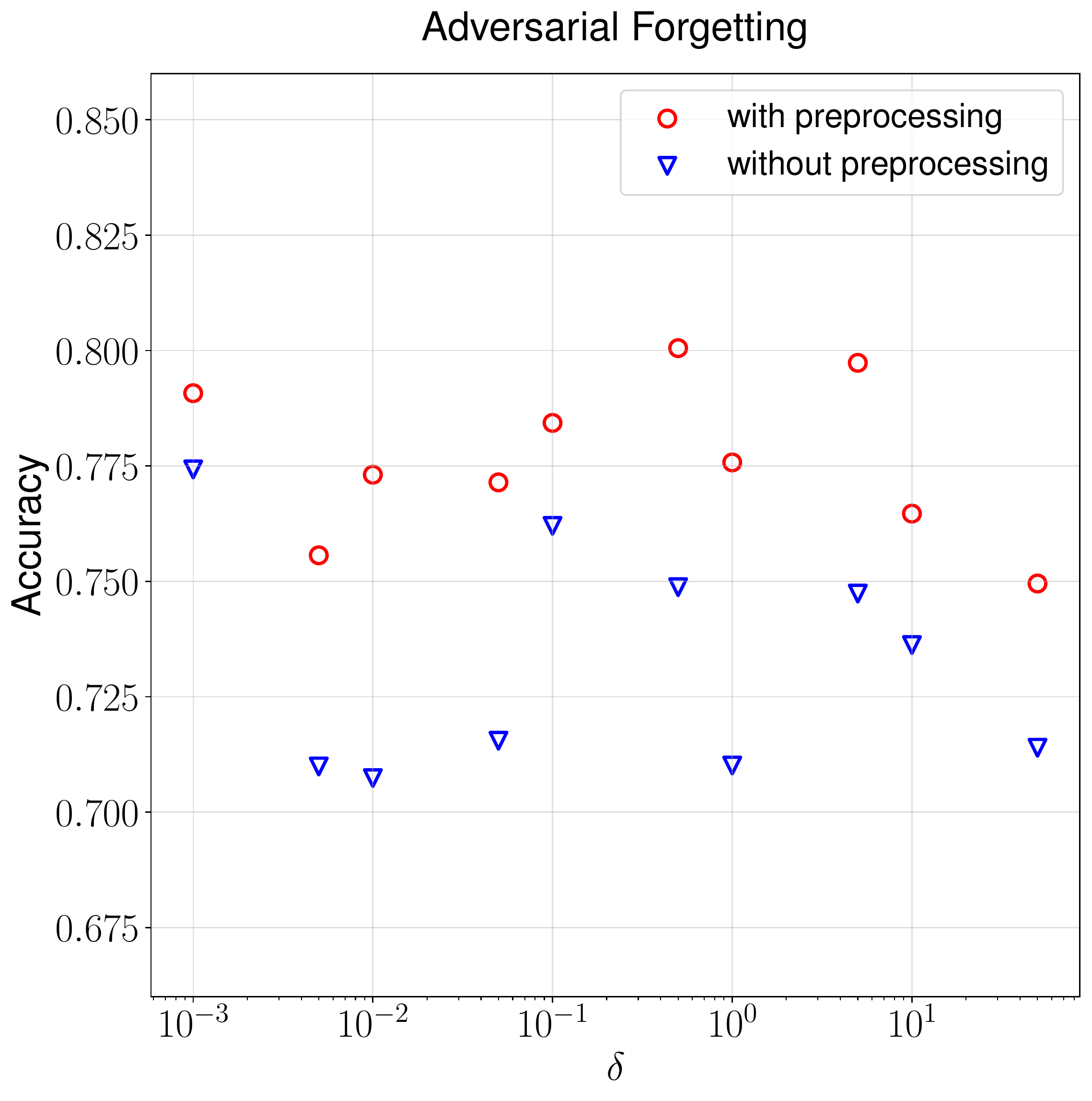}
        \caption{Accuracy of predicting $c$ from $z$ with and without pre-processing as a function of discriminator coefficient $\delta$. For this, we fixed $\rho=0.001$ and $\lambda=0.1$.}\label{fig:invariance_adv_forget1}
    \end{subfigure}
    \quad
    \begin{subfigure}[]{0.47\textwidth}
        \includegraphics[width=\textwidth]{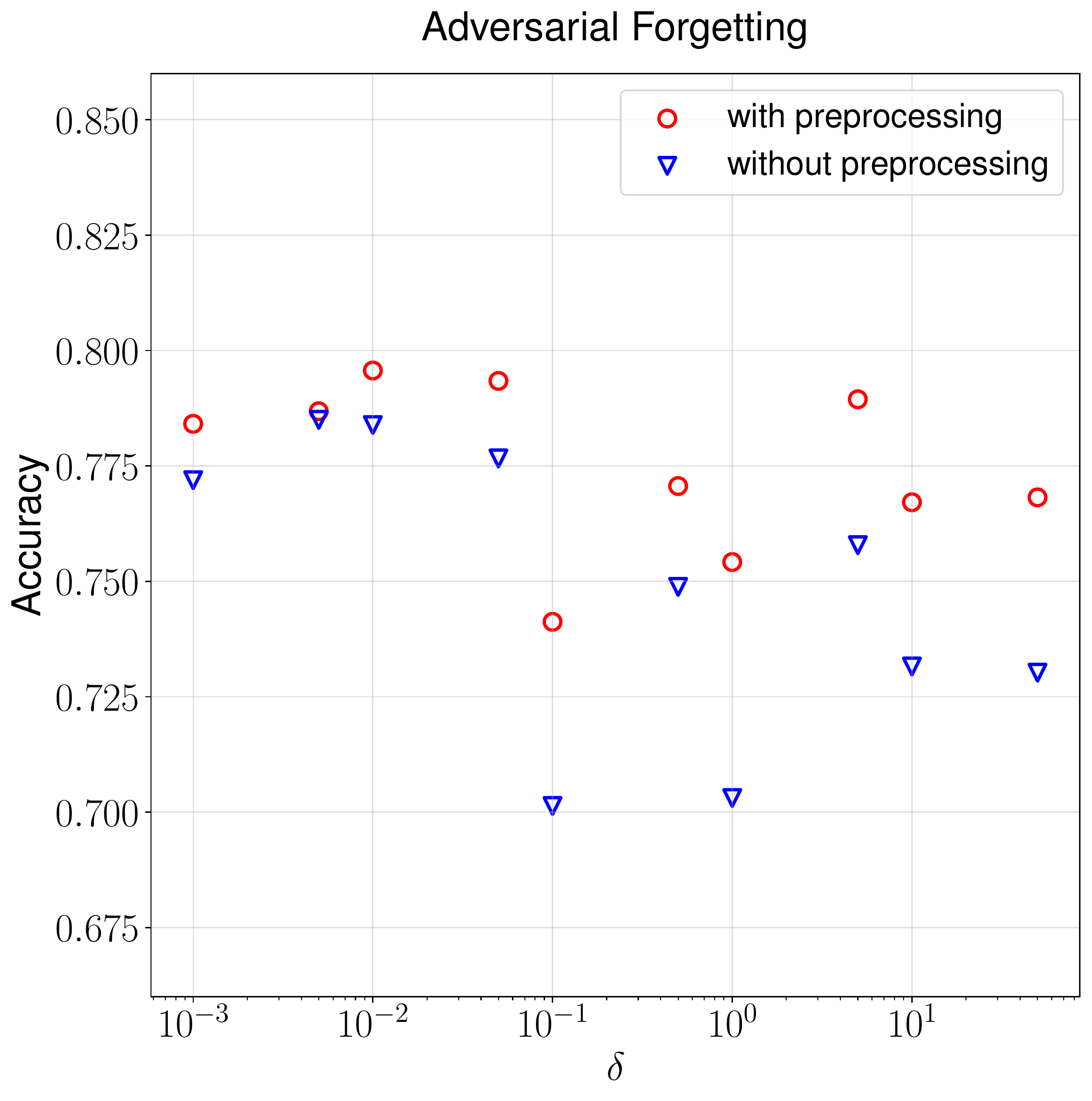}
        \caption{Accuracy of predicting $c$ from $z$ with and without pre-processing as a function of discriminator coefficient $\delta$. For this, we fixed $\rho=0.001$ and $\lambda=0.01$.}\label{fig:invariance_adv_forget2}
    \end{subfigure}
    \caption{Predicting $c$ from representations provided by MaxEnt-ARL and Adversarial Forgetting on \textit{UCI Adult} dataset. Majority baseline is around 0.68. We show two representative plots for adversarial forgetting from the 9 possible values of $(\lambda,\rho)$.}
\end{figure}

Adversarial representation learning is shown to be useful for learning fair representations by enforcing invariance. Invariance is the strongest requirement for fairness. If the representations are invariant, i.e.,  $I(\mbz:\mbc)=0$, then representations are perfectly fair and  achieve $\DP=0$. However, in our experiments, adversarial methods of \citet{jaiswal2020invariant, roy2019mitigating} could not achieve lower \DP\ or only did at excessive loss of utility, which conflicts with their results.

\citet{jaiswal2020invariant, roy2019mitigating} reported near-perfect invariance with a negligible drop in predictive performance for the \textit{UCI Adult} dataset. They demonstrated invariance by showing that a classifier trained to predict $c$ from $z$ performs poorly.
We have  shown that this is only a lower bound for $I(\mbz:\mbc)$ (via Eq.~\ref{eq:lower_bound_classifier}) and, therefore, cannot be used to claim invariance or fairness with any confidence. A better model or a different prediction technique may still be able to predict well. \citet{xu2020theory} (Sec 6.4 and D.2 in their paper) has also discussed this. They have shown that if the classifier (which predicts $c$ from $z$) is different from the adversary used during the training, there is no guarantee about the invariance or fairness of representations. Here, we investigated the discrepancy in our results of not achieving lower \DP . Our experiments reveal that it is possible to predict sensitive information $c$  even with the same model as the adversary.

Figs.~\ref{fig:invariance_maxent},~\ref{fig:invariance_adv_forget1}, and~\ref{fig:invariance_adv_forget2} show results of predicting $c$ from $z$ for different values of loss coefficient for MaxEnt-ARL and Adversarial Forgetting for the \textit{UCI Adult} dataset. We use the same classifier architecture as the  adversary used during training (See Table~\ref{tab:arch_hidden_layer_size}).   Using simple pre-processing step like standard scaling (using only train set statistics), we could predict $c$ from $z$; however, by training the same classifier on the representations without pre-processing, we were not able to predict $c$ at all.

One might suspect that this issue can be alleviated using batch-norm in the discriminator or adversary during the training as the batch-norm operations are similar to standard scaling.  Fig.~\ref{fig:invariance_maxent_bn} shows the results of using batch-norm in the discriminator. We use the batch-norm layer after input as well as the hidden layer. With or without batch-norm, our observations do not change much. This suggests that adversarial representation learning methods may not remove information from the representation and only obfuscate it such that the adversary can not  see it.

\section{Area Over Parity-Accuracy Curve}
\label{sec:aopac}
\begin{table}
	\centering
	\begin{tabular}{l ccccc ccccc}
		\toprule
		\multirow{2}{*}{Method}
		& \multicolumn{4}{c}{UCI Adult} &  \multicolumn{4}{c}{Heritage Health}\\
		\cmidrule(lr){2-6} \cmidrule(lr){7-11}

		& LR & SVM &  MLP (2) & MLP (1) & RF & LR & SVM &  MLP(2) & MLP(1) & RF\\
		\cmidrule(l){1-1} \cmidrule(lr){2-2} \cmidrule(lr){3-3} \cmidrule(lr){4-4} \cmidrule(lr){5-5} \cmidrule(lr){6-6} \cmidrule(lr){7-7} \cmidrule(lr){8-8} \cmidrule(lr){9-9}  \cmidrule(lr){10-10}  \cmidrule(lr){11-11}
		FCRL (Ours) 			& 0.309 & 0.327 & 0.246 & 0.307 & 0.303   & 0.294 & 0.322 &0.287  & 0.319& 0.310 \\
		CVIB 					& 0.192 & 0.218 & 0.105 & 0.182 & 0.166   & 0.202 & 0.234 &0.179  & 0.191& 0.193 \\
		MIFR 					& 0.221 & 0.249 & 0.189 & 0.251 & 0.231   & 0.193 & 0.222 &0.167  & 0.202& 0.196 \\
		MaxEnt-ARL 				& 0.180 & 0.205 & 0.122 & 0.144 & 0.110   & 0.0   & 0     &0      & 0.0	 & 0.015 \\
		LAFTR 					& 0.235 & 0.264 & 0.105 & 0.155 & 0.089   & N/A   & N/A   &N/A    & N/A  & N/A   \\
		Adversarial Forgetting 	& 0.092 & 0.114 & 0.070 & 0.087 & 0.025   & 0.173 & 0.185 &0.168  & 0.156& 0.067 \\
		\bottomrule
	\end{tabular}
	\caption{Area Over Parity-Accuracy Curve with different downstream classification algorithms (Higher is better). The normalization area is computed using LP (Eq.~\ref{eq:LP_frontier}).
	LR stands for Logistic Regression, RF stands for random forest and MLP(x) is a x hidden layer MLP with 50 hidden units in each layer.
	}
	\label{tab:AOPAC_better}
\end{table}

\begin{figure}[]
    \centering
    \begin{subfigure}[]{\textwidth}
        \centering
        \includegraphics[width=\textwidth]{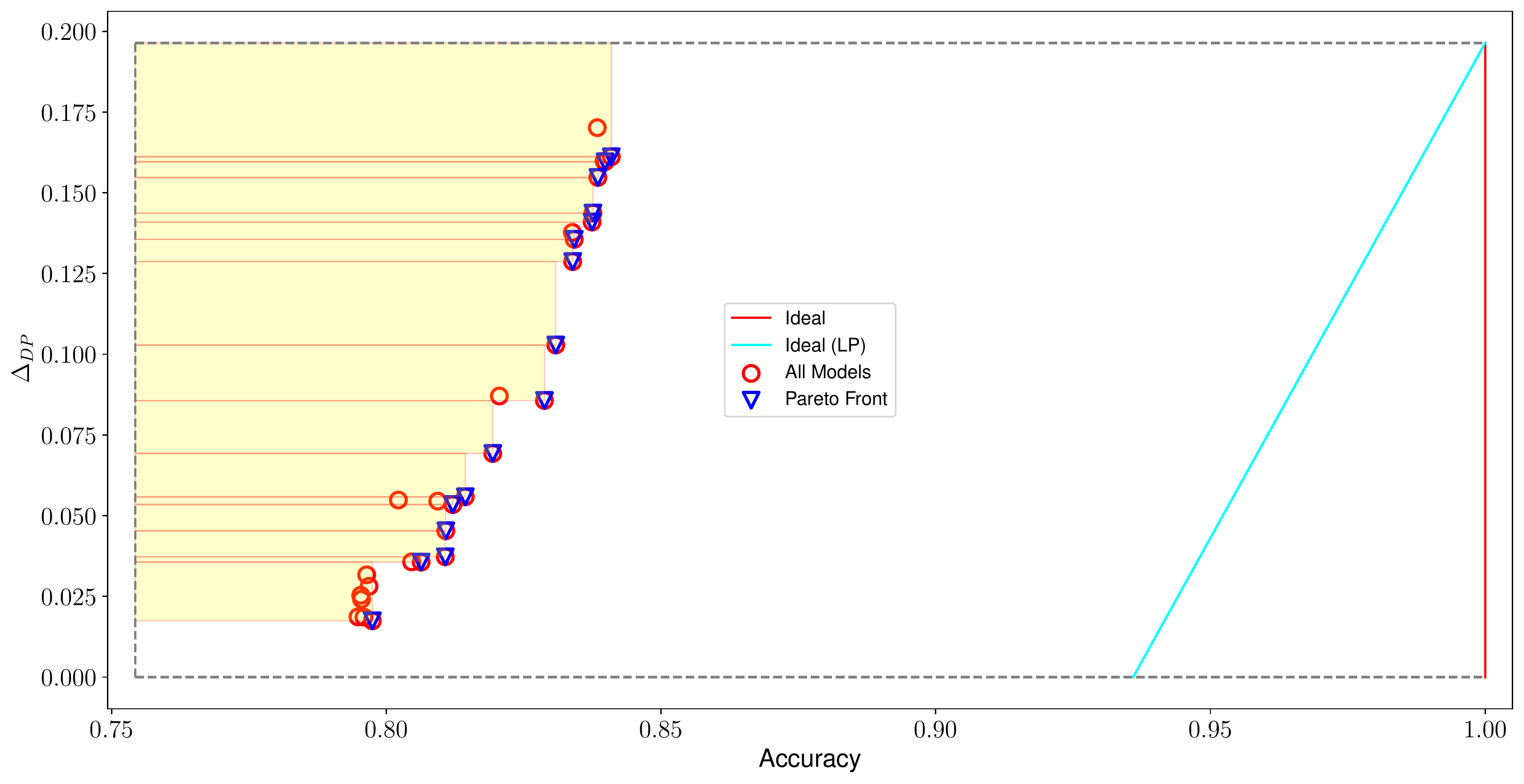}
        \caption{\textit{UCI Adult} dataset.} %
    \end{subfigure}

    \vspace{30pt}
    \begin{subfigure}[]{\textwidth}
        \centering
        \includegraphics[width=\textwidth]{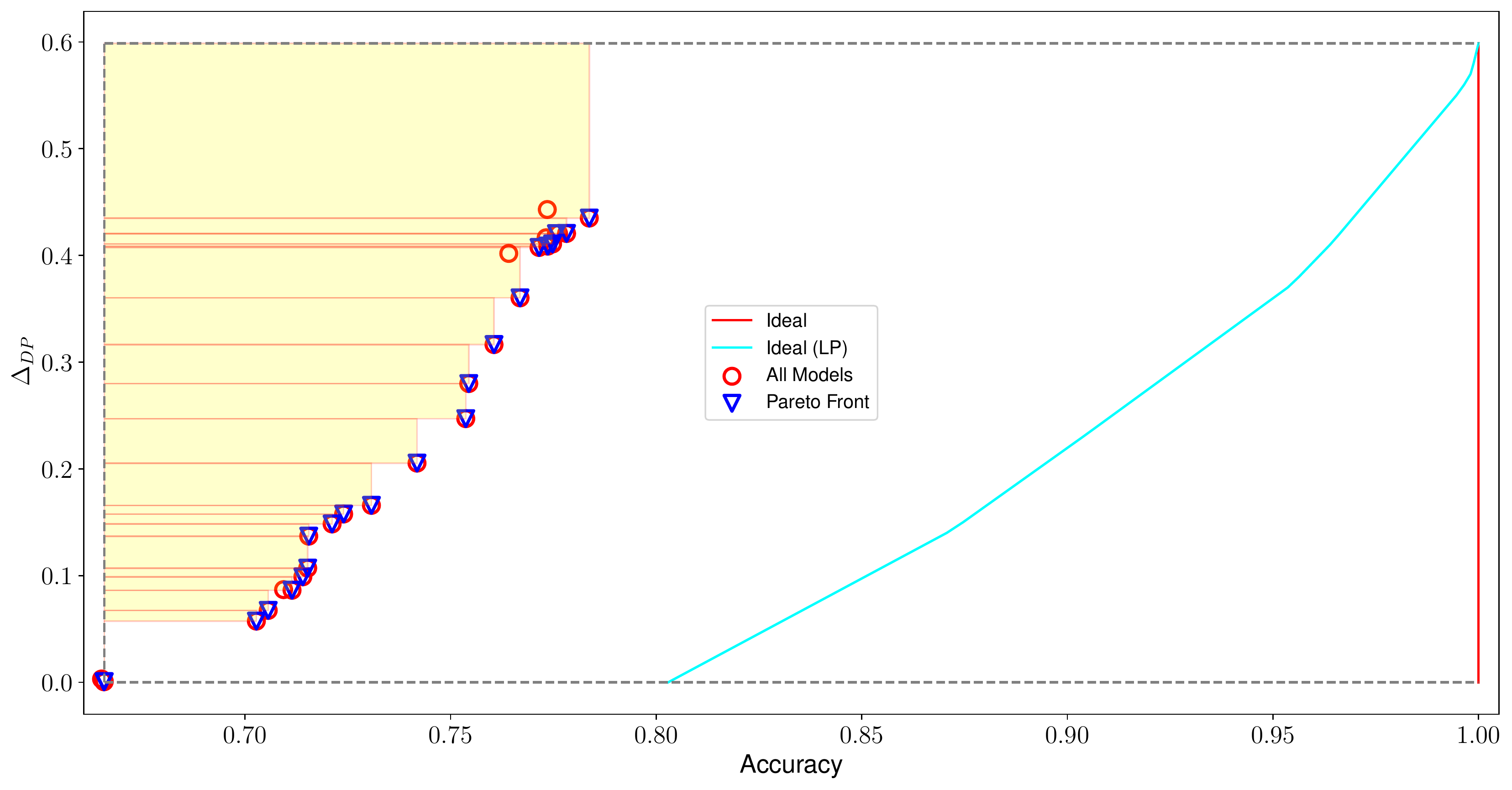}
        \caption{\textit{Heritage Health} dataset.}%
    \end{subfigure}
    \caption{
        Visualizing area over the parity-accuracy curve. The points shown are from FCRL. Pareto optimal points are in blue. We may normalize with the area spanned by the region within dashed lines and the red line. However, a better bound can be found by solving LP in Eq.~\ref{eq:LP_frontier}, which is shown in cyan. Area over curve results that are normalized by the area from LP solutions are   in Table~\ref{tab:AOPAC_nn_1_layer} and~\ref{tab:AOPAC_better}.}
    \label{fig:area_over_curve_demo}
\end{figure}

The parity-accuracy curve provides  a visual evaluation of different methods. Any method that pushes the achievable frontier to the right and bottom is preferable. A practitioner can use parity-accuracy plots to find the most predictive representations under prescribed parity thresholds or vice-versa.
A more efficient method must be able to provide more predictive representation at the same parity. To this end, we state the following desiderata
from a fair representation learning method:
\begin{itemize}
    \item The method should  provide representations with high utility for any reasonable parity constraints.
    \item The method should  provide representations for all feasible  parities\footnote{Some methods might not achieve low \DP. In such cases, the parity can be satisfied  using random representations that would give 0 \DP\ and no utility. For example: Consider Fig.~\ref{fig:area_over_curve_demo}, when very low \DP\ like 0.001 is desired, no representation can be chosen from the method.}.
\end{itemize}
\noindent
\noindent
In Fig.~\ref{fig:area_over_curve_demo}, we  see that the yellow region is the feasible region, i.e., if the prescribed constraints are within this region, the method can provide representations that satisfy the constraints. The area is precisely the area over parity-accuracy curve. Therefore, any method that covers more area is superior. Using this as the motivation, we compute the area over the parity-accuracy curve to provide a quantitative evaluation. It admits an intuitive interpretation as the area of the feasible region of parity and accuracy, i.e., if we choose any point in this region, the method can provide representations with a better trade-off. Next, we discuss finer details for computing this area.

\paragraph{Filtering Points and Finding Feasible Region:}
An ideal parity-accuracy curve would be a smooth, increasing line. However, due to practical constraints, we can only draw finite points, and  due to the inherent randomness of the training procedure,  we might not get an exact monotonically increasing behavior. Therefore, the first step in computing this metric is to compute a Pareto front, and we discard any point with higher \DP\ and lower accuracy than some other point. We also discard representations with \DP\ more than \DP\ of data (computed using the true labels) since we want to achieve fairness and are only interested in representations with lesser \DP. We compute the area using the remaining points. Since we have finite points, if the two adjacent points are $(u_1, p_1)$ and $(u_2, p_2)$, $p_1<p_2$.  $p,u $ denote parity and utility, respectively. For the width of the bar between $p_1$ and $p_2$, we use $u_1$ as that is the achievable accuracy. Accuracy associated with $p_2$ is only achievable when the parity desired is more than $p_2$.

\paragraph{Normalization:}
While the area gives us a relative idea of the method's performance, it would be ideal to have a metric where the maximum value is not dataset dependent. Therefore, we also normalize it by the maximum possible achievable area. To normalize the area, we next consider the limits on parity and accuracy that a model can achieve.

\paragraph{Limits on \DP\ and Accuracy:}
The theoretical extremum of both \DP\ and accuracy is 1. Nevertheless, since  we are trying to reduce the \DP\ and therefore, a more realistic limit of \DP\ would be the \DP\ computed from the test/validation set. Let us call this value $\DP^{(data)}$. We would like fair representation learning models to have \DP\ lower than this. Therefore we discard any representations that may have \DP\ more than $\DP^{(data)}$. Bayes-Optimal Classifier would give the maximum accuracy that can be achieved, but it is not realistically computable. Therefore, the upper bound of 1 on accuracy is a reasonable choice. We may use some other feasible model, but it makes the metric rely on the model choice. Therefore, we avoid this choice and use  $(1,\DP^{(data)})$ as one extremum of the achievable frontier. For the other extremum, we may use $(1,0)$, i.e., 0 \DP\ and perfect accuracy. However, we can compute a more realistic frontier by solving simple linear programs.

\paragraph{A More Realistic Optimal Frontier:}
Since $I(\mby:\mbc)>0$, we know that if representations are to be fair, predictions $\hat y$ will be affected. We can take this into account and  can further bound the achievable frontier. We assume that the optimal classifier knows the correct labels of the test set, i.e., it is perfectly accurate. To achieve the desired \DP\ constraints, while minimizing error, we can adjust the decisions so that some of the labels in each $c=i$ group are flipped. We can write this as a linear program as there are only two classes ($y\in \{0,1\}$).

\begin{align}
    \min \sum_i & |\delta_i|  \ \ \ \ \text{such that} \nonumber \\
    &\left  | \cfrac {P(\mby=1, \mbc=i)-\delta_i}{P(\mbc=i)}  -\cfrac {P(\mby=1, \mbc=j)-\delta_j}{P(\mbc=j)} \right| \leq  \Delta &\forall i,j \nonumber \\
    & -P(\mby=0, \mbc=i) \leq \delta_i \leq P(\mby=1, \mbc=i)  &\forall i\label{eq:LP_frontier}
\end{align}
where $\delta_i$ denotes the probability mass of group $i$ whose labels are flipped from 1 to 0, if it is negative, that means 0 labels are flipped to 1. $\sum_i |\delta_i|$ will be the error. $\Delta$ denotes the desired \DP\ and the first constraint will enforce that. Lastly, we cannot flip more probability mass than available for each group. We solve this linear program for different values of $\Delta$.

\paragraph{Steps Involved:}
We subtract the random or majority baseline from the accuracy. The following steps  are involved in computing this metric, and  the visualizations are shown in Fig.~\ref{fig:area_over_curve_demo}.
\begin{enumerate}
    \item Compute the Ideal area from the data and $\DP^{(data)}$. This will be used for normalization.
    \item Discard parity-accuracy pairs that have \DP\ higher than $\DP^{(data)}$.
    \item Compute non-dominated parity-accuracy pairs, i.e., compute the Pareto front.
    \item Draw bars between adjacent parity with width as the accuracy of the lower parity point.
    \item Compute the normalized area of the bars.
\end{enumerate}

\fi
\end{document}